\documentclass{article}

\usepackage{PRIMEarxiv}

\usepackage[utf8]{inputenc} 
\usepackage[T1]{fontenc}    

\usepackage{nicefrac}       
\usepackage{microtype}      
\usepackage{lipsum}
\usepackage{fancyhdr}       
\usepackage{amsmath,amssymb,amsfonts}
\usepackage{textcomp}
\usepackage{xcolor}
\usepackage{tipa}
\usepackage{rotating}
\usepackage{booktabs}
\usepackage{subcaption}
\usepackage{lineno}
\usepackage{soul}
\usepackage{multirow}
\usepackage{hyperref}
\usepackage{tablefootnote}
\usepackage{xurl}

\usepackage{algorithm}%
\usepackage{algorithmicx}%
\usepackage{algpseudocode}%
\usepackage{authblk}

\usepackage{graphicx}
\graphicspath{{Figures/}}

\title{Training and Agentic Inference Strategies for LLM-based Manim Animation Generation
}

\author[1]{\textbf{Ravidu Suien Rammuni Silva}}
\author[1]{\textbf{Ahmad Lotfi}}
\author[1]{\textbf{Isibor Kennedy Ihianle}}
\author[2]{\textbf{Golnaz Shahtahmassebi}}
\author[1]{\\\textbf{Jordan J. Bird}}

\affil[1]{Department of Computer Science, Nottingham Trent University, Nottingham, United Kingdom\\
\texttt{\{ravidu.rammunisilva, ahmad.lotfi, isibor.ihianle, jordan.bird\}@ntu.ac.uk}}

\affil[2]{Department of Physics and Mathematics, Nottingham Trent University, Nottingham, United Kingdom\\
\texttt{golnaz.shahtahmassebi@ntu.ac.uk}}

\begin{document}
\maketitle

\begin{abstract}
Generating programmatic animation using libraries such as Manim presents unique challenges for Large Language Models (LLMs), requiring spatial reasoning, temporal sequencing, and familiarity with domain-specific APIs that are underrepresented in general pre-training data.
A systematic study of how training and inference strategies interact in this setting is lacking in current research. This study introduces ManimTrainer, a training pipeline that combines Supervised Fine-tuning (SFT) with Reinforcement Learning (RL) based Group Relative Policy Optimisation (GRPO) using a unified reward signal that fuses code and visual assessment signals, and ManimAgent, an inference pipeline featuring Renderer-in-the-loop (RITL) and API documentation-augmented RITL (RITL-DOC) strategies. Using these techniques, this study presents the first unified training and inference study for text-to-code-to-video transformation with Manim.
It evaluates 17 open-source sub-30B LLMs across nine combinations of training and inference strategies using ManimBench. Results show that SFT generally improves code quality, while GRPO enhances visual outputs and increases the models' responsiveness to extrinsic signals during self-correction at inference time. 
The Qwen 3 Coder 30B model with GRPO and RITL-DOC achieved the highest overall performance, with a 94\% Render Success Rate (RSR) and 85.7\% Visual Similarity (VS) to reference videos, surpassing the baseline GPT-4.1 model by +3 percentage points in VS. Additionally, the analysis shows that the correlation between code and visual metrics strengthens with SFT and GRPO but weakens with inference-time enhancements, highlighting the complementary roles of training and agentic inference strategies in Manim animation generation.
\end{abstract}



\section{Introduction}
Programmatically generated animations show promise as reliable mediators for the efficient, automated generation of educational videos using Large Language Models (LLMs). Leveraging LLMs' code-generation abilities with simple yet effective animation libraries, such as Manim \cite{manim_community_2026}, provides greater accuracy and efficiency than diffusion-based video generation models. It is efficient, as video generation pipelines rely on diffusion-based models that require significantly more computing resources than compact LLMs, and it is accurate, since the text-to-code-to-video process makes the generated video precise by relying on the generated code, especially in mathematical animations such as graphing.
LLMs have demonstrated increasingly accurate code-generation capabilities \cite{chen_evaluating_2021, hui_qwen25-coder_2024, seed_seed-coder_2025}. However, generating correct Manim code remains difficult, as it requires not only syntactic fluency in Python but also reasoning about three-dimensional coordinate systems, sequencing of animations over time, and knowledge of the Manim API, which is under-represented in general code-based pre-training datasets \cite{silva_large_2026}.

Recent advancements in Parameter-efficient Fine-tuning (PEFT), notably LoRA \cite{hu_lora_2021} and the more efficient quantised variant QLoRA \cite{dettmers_qlora_2023}, have enabled specialising sub-30B LLMs on a single consumer-grade GPU. Supervised Fine-tuning (SFT) can establish a lexical foundation for the Manim API, while Reinforcement Learning (RL) based approaches, such as Group Relative Policy Optimisation (GRPO) \cite{shao_deepseekmath_2024, silva_large_2026}, can further improve the model through execution-based reward signals without requiring a separate critic model. SFT, together with RL, encompasses the text-to-code-to-video process in LLM-based Manim animation generation. However, the interaction between these distinct training strategies and how each contributes to the code quality of the generated Manim code, as well as the visual quality of the video it renders, has not been systematically examined.

At inference time, iterative self-correction with extrinsic feedback from compilers has been shown to effectively improve the quality of LLM-generated code \cite{chen_teaching_2024}. Additionally, Retrieval-Augmented Generation (RAG) has been demonstrated to decrease API hallucinations by incorporating relevant API documentation into the prompt context \cite{lewis_retrieval-augmented_2020, wang_coderag-bench_2024}. However, the interaction between the training strategies mentioned earlier and these inference-time strategies remains unexplored, particularly in domains where both the generated code and its visual output need to be assessed.

This need for a multifaceted evaluation requires metrics that span both the code and visual domains and encompass the entire text-to-code-to-video process. Code metrics such as CodeBLEU \cite{ren_codebleu_2020} and CodeBERT similarity \cite{feng_codebert_2020} may focus solely on the generated code, while visual metrics such as SSIM \cite{ndajah_ssim_2010} and CLIP-based vector similarity \cite{radford_learning_2021} can assess visual quality and accuracy but overlook the code quality. Moreover, the extent to which these two metric families agree across different training and inference strategies has not yet been studied.

This study hypothesises that combining supervised fine-tuning (SFT) and GRPO-based reinforcement learning with agentic inference strategies would yield better performance in the LLM-based text-to-code-to-video process.

Based on the identified gaps and the formulated hypothesis above, this work makes four key contributions: 
\begin{enumerate}
    \item \textbf{ManimTrainer}: A training pipeline that combines SFT with GRPO using a unified reward function that fuses code metrics with visual metrics to fine-tune LLMs for the Manim code generation task, covering the full text-to-code-to-video process.
    \item \textbf{ManimAgent}: An inference pipeline featuring Renderer-in-the-loop (RITL) self-correction and API documentation-augmented RITL (RITL-DOC) inference strategies to further improve the performance of LLMs in Manim code generation at inference time.
    \item A comprehensive evaluation of 17 open-source sub-30B LLMs ranging from 0.5B in size using the unified ManimTrainer and ManimAgent pipeline, examining the individual effects and interactions between training-time optimisation (SFT, GRPO) and agentic inference strategies (RITL, RITL-DOC) using both code and visual metrics.
    \item An analysis of the correlation between code and visual metrics across all strategy combinations, showing that inference-time strategies weaken their correlation while improving resultant visual quality, highlighting limitations of code-only evaluation.
\end{enumerate}

The remainder of this paper is organised as follows. Section~\ref{sec:related-work} reviews related work. Section~\ref{sec:methodology} presents the technical details of ManimTrainer and ManimAgent. Section~\ref{sec:results} presents the results of the experiments, discusses the findings and limitations, and Section~\ref{sec:conclusion} concludes the paper.

\section{Related Work}
\label{sec:related-work}

\subsection{SFT and RL for LLM-based Code Generation}
LLM-based code generation has progressed rapidly from Codex \cite{chen_evaluating_2021} to recent compact models such as Qwen 2.5 Coder \cite{hui_qwen25-coder_2024} and SeedCoder \cite{seed_seed-coder_2025}, which rival much larger systems on standard benchmarks \cite{zhuo_bigcodebench_2024}. However, these models often underperform on domain-specific libraries with custom APIs \cite{sun_januscoder_2025, silva_large_2026}.

While SFT establishes vocabulary and syntactic foundations for code generation, it does not incorporate execution-based feedback \cite{guo_deepseek-r1_2025}. Reinforcement learning addresses this gap, but earlier approaches such as PPO \cite{ouyang_training_2022} and DPO \cite{rafailov_direct_2023} require a separate critic network or preference pairs.

Group Relative Policy Optimisation (GRPO) \cite{shao_deepseekmath_2024} addresses this by offering a critic-free alternative suited to code generation, where outputs are evaluated using non-differentiable execution-based signals. While DeepSeek-R1 demonstrated that GRPO can significantly improve functional correctness \cite{guo_deepseek-r1_2025}, the interaction between SFT- and GRPO-trained models with inference-time enhancement strategies remains unexplored.

\subsection{Inference-time enhancements and Self-correction}
Iterative self-correction approaches such as Self-Debugging \cite{chen_teaching_2024}, Reflexion \cite{shinn_reflexion_2023}, and Self-Refine \cite{madaan_self-refine_2023} allow LLMs to improve code via in-context learning (ICL), though systematic evaluations show intrinsic correction yields only marginal gains \cite{kamoi_when_2024} and extrinsic validation is significantly more effective \cite{tie_can_2025}.

Extrinsic feedback from compilers, interpreters and renderers can significantly improve the self-correction capabilities of LLMs. CodeRL \cite{le_coderl_2022}, InterCode \cite{yang_intercode_2023}, and SWE-Agent \cite{yang_swe-agent_2024} demonstrate that more informative error messages lead to faster convergence \cite{chen_teaching_2024}. ReLook \cite{li_relook_2025} extends this to the visual domain, using a multimodal LLM-based critic to detect defects that execution logs cannot surface.

Supplementing the self-correction loops, Retrieval-Augmented Generation (RAG) has the potential to address text-based errors, such as hallucinated APIs and their parameters, by injecting relevant documentation into the prompt context \cite{lewis_retrieval-augmented_2020}. While RAG consistently reduces hallucinations, models can struggle with dense technical APIs and longer contexts \cite{wang_coderag-bench_2024}. The interaction between API information augmentation with iterative execution and/or renderer-based refinement loops, however, remains largely unexplored.

\subsection{Programmatic Animation Generation via LLMs}
Combined with the code generation capabilities of LLMs, libraries such as Manim \cite{manim_community_2026} offer a more efficient and precise alternative to generating animations with LLMs. While other works generate visuals via HTML or SVG code \cite{tseng_keyframer_2024, zhao_vincicoder_2025, liu_logomotion_2025}, Manim enables complex animations in a few lines of Python. However, it requires spatial reasoning about 3D coordinates, temporal sequencing of animations, and knowledge of a specialised API that is not well represented in general pre-training corpora.

ManimBench \cite{silva_large_2026} addresses the need for a domain-specific evaluation benchmark for Manim by providing a dataset of 417 human-reviewed, paired natural-language descriptions and executable Manim code snippets. Automated pipelines such as Manimator \cite{p_manimator_2025} and TheoremExplainAgent \cite{ku_theoremexplainagent_2025} employ multi-stage processes to generate code for visualising educational and academic content, primarily using commercial, API-based, closed-source LLMs. Code2Video \cite{chen_code2video_2025} further explores this multi-stage process, highlighting the gap between the code execution success and the quality of the visual outputs.

\subsection{Evaluating LLM-generated Code and its Visual Output}
Code-level metrics such as CodeBLEU \cite{ren_codebleu_2020} and CodeBERT similarity \cite{feng_codebert_2020} evaluate the structure of generated code, but may penalise Manim snippets that produce similar visual outputs via different coding pathways \cite{silva_large_2026}.

Visual metrics such as Structural Similarity Index (SSIM) \cite{wang_image_2004, ndajah_ssim_2010} and Contrastive Language-Image Pretraining (CLIP) \cite{radford_learning_2021} based schematic similarity can evaluate the quality of rendered outputs against reference frames, capturing visual fidelity that code-level metrics overlook. Conversely, relying solely on visual metrics may reward inefficient code. This necessitates combining both metric families and extending evaluation to the temporal dimension of video.

\section{Methodology}
\label{sec:methodology}
\begin{figure}[t]
    \centering
    \includegraphics[width=1\linewidth]{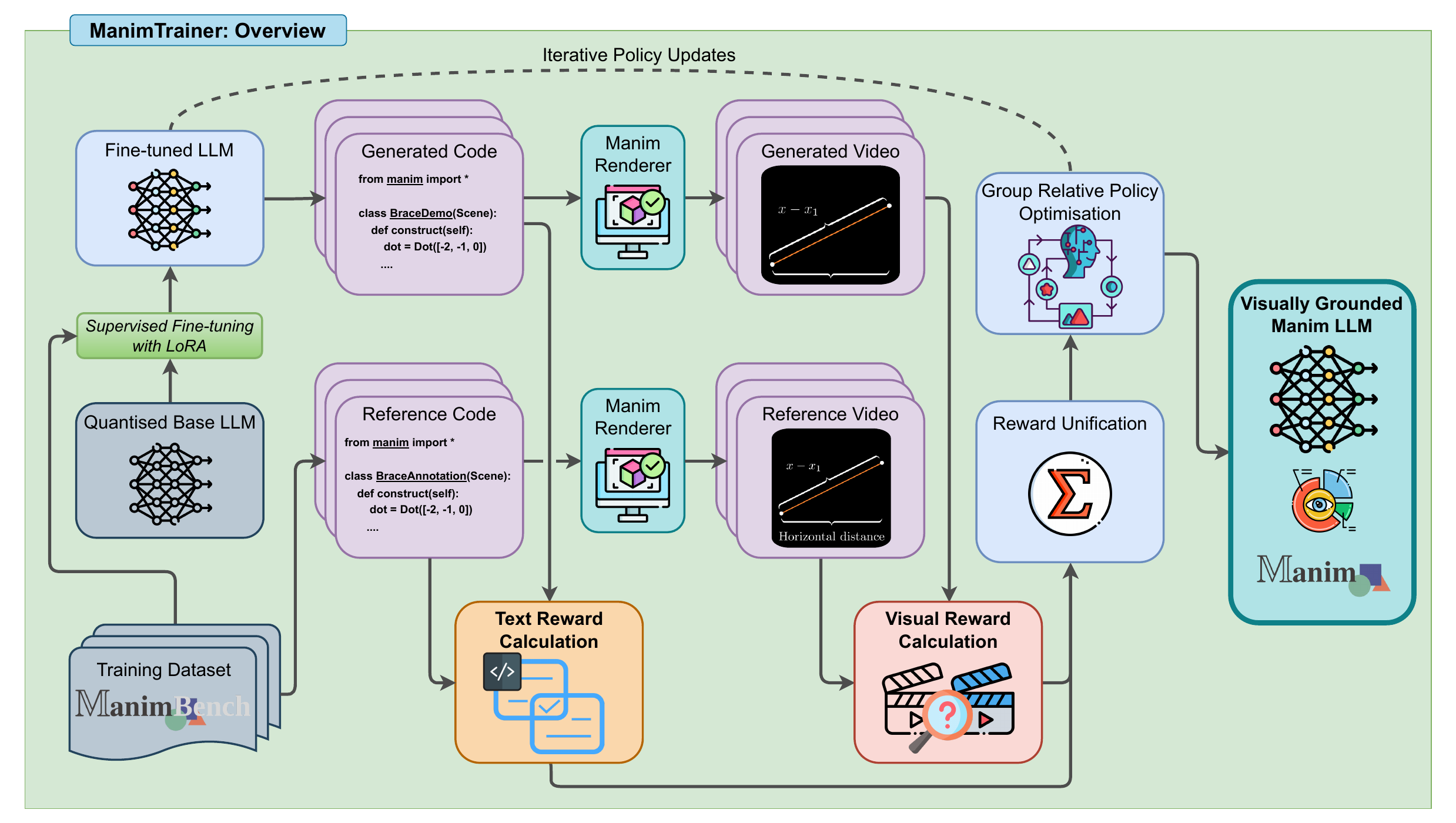}
    \caption{Overview of the ManimTrainer pipeline. The pipeline uses a quantised base LLM and visually grounds it for Manim code generation, using both visual and text reward signals.}
    \label{fig:overview}
\end{figure}

This section presents the training pipeline, ManimTrainer, and the inference pipeline, ManimAgent, used in this study to fine-tune and generate Manim code using sub-30B LLMs. The methodology comprises four key stages. First, SFT provides the key understanding of the task and adds Manim-specific knowledge to the LLM at a parametric level. Second, a unified reward function $\mathcal{R}$ is formulated by fusing a code-based text reward $\mathcal{R_T}$ with a visual reward $\mathcal{R_V}$ computed from the visual similarity between the generated video and the reference video. Third, the formulated reward function is utilised to fine-tune the LLM via RL, using GRPO to visually ground the SFT-trained LLM. Fourth, inference-time enhancements, Renderer-in-the-loop (RITL) and Manim API documentation-supported RITL (RITL-DOC), are introduced to be applied to the fine-tuned models. Together, these stages support the systematic investigation of this study's hypothesis. An overview of the ManimTrainer pipeline is illustrated in Figure~\ref{fig:overview}, while the strategised inference pipeline is summarised in Figure~\ref{fig:inference-pipeline}. The following subsections detail each stage.

\subsection{SFT Optimisation}
Using the ManimBench dataset \cite{silva_large_2026}, the models are first trained with SFT to establish Manim-specific vocabulary and syntax. The detailed SFT and GRPO training configurations are presented in Section~\ref{subsubsec:training-setup}. 

\subsection{Reward Function}
\label{subsec:reward-function}

Let a training instance $(x,\hat{y})$, where $x$ is the formulated prompt which includes the textual description and $\hat{y}$ is the reference Manim code snippet. During the RL training cycle using GRPO, policy $\pi_\theta$ samples completion $c \sim \pi_\theta(\cdot|x)$ from which executable Manim code is deterministically extracted using $\mathcal{H}$:

\begin{equation}
\label{eq:code-extraction}
    y = \mathcal{H}(c) ,
\end{equation}

\noindent where $\mathcal{H}$ is a regex-based function that extracts the generated code snippet $y$, where $\mathcal{H}(c)=\varnothing$ when no code can be extracted. The regex function attempts to find the code wrapped between \texttt{<CODE>} and \texttt{</CODE>} tags and/or \texttt{\textasciigrave\textasciigrave\textasciigrave python} markdown-based fences. The extracted Manim code $y$ is then used to calculate the Text Reward $\mathcal{R_T}$ and the Visual Reward $\mathcal{R_V}$, and to derive the Unified Reward $\mathcal{R}$ as shown below.

\subsubsection{Text Reward (\texorpdfstring{$\mathcal{R_T}$)}{Rt}}
\label{subsubsec:text-reward-calc}
The text reward is based on two code-oriented similarity metrics, CodeBLEU ($\operatorname{C-BLEU}$) \cite{ren_codebleu_2020} and CodeBERT similarity ($\operatorname{C-BERT}$) \cite{feng_codebert_2020}. $\operatorname{C-BERT}$ is calculated as below, taking the cosine similarity of the two embeddings obtained for the reference code and the generated code:

\begin{equation}
    \operatorname{C-BERT} = \cos(f_{\text{C-BERT}}(y),f_{\text{C-BERT}}(\hat{y})) ,
\end{equation}

\noindent where $f_{\text{C-BERT}}(y)$ denotes the mean-pooled final hidden layer of embedding $y$. Then, finally, the text reward $\mathcal{R_T}$ is calculated by obtaining the geometric mean of the two scores as below:

\begin{equation}
    \mathcal{R_T}(y,\hat{y}) = \sqrt{\operatorname{C-BLEU}(y,\hat{y}) \cdot \operatorname{C-BERT}(y,\hat{y})}
    .
\end{equation}

\subsubsection{Visual Reward (\texorpdfstring{$\mathcal{R_V}$)}{Rv}}
\label{subsubsec:visual-reward-calc}
Firstly, rendered videos $v$ and $\hat{v}$ are produced:

\begin{equation}
\label{eq:manim-renderer}
    v = \mathcal{M}(y), \hat{v} = \mathcal{M}(\hat{y}) ,
\end{equation}

\noindent where $\mathcal{M}$ represents the Manim rendering engine.

Then, to ensure visual correctness, as indicated by visual similarity to the expected video, the rendered videos $\hat{v}$ and $v$ are compared using both structural similarity $\mathcal{S}_{ssim}$ and semantic similarity $\mathcal{S}_{sem}$ measures with dynamic time warping ($\operatorname{DTW}$) alignment to fairly compare two videos of different lengths. The comparison is performed at each frame, with frames sampled at a fixed rate of 5  frames per second (FPS). The sample rate was set empirically, based on many Manim-based GIF images that work at 5 FPS.

From each video pair ($v$,$\hat{v}$), a pair of sets of grayscale frames ($G$,$\hat{G}$) (for $\mathcal{S}_{ssim}(v,\hat{v})$ calculation) and a pair of sets of RGB frames ($I$,$\hat{I}$) (for $\mathcal{S}_{sem}(v,\hat{v})$ calculation) are obtained.

Firstly, the SSIM matrix between all elements of $G$ and $\hat{G}$ is computed:
\begin{equation}
    S^\prime = \operatorname{SSIM}(G,\hat{G}), \; S^\prime \in \mathbb{R}_{T\times\hat{T}},
\end{equation}

\noindent where $T$ and $\hat{T}$ represents frame counts sampled from respectively $v$ and $\hat{v}$.

Then the matrix $S^\prime$ is summarised into two sequences, $S^\prime_{ref}$ and $S^\prime_{gen}$, yielding per-frame best matches for the reference and generated videos, respectively:

\begin{equation}
    S^\prime_{ref} = \max_T S^\prime, S^\prime_{gen} = \max_{\hat{T}} S^\prime ,
\end{equation}

Then, $S^\prime_{ref}$ and $S^\prime_{gen}$ are aligned using $\operatorname{DTW_{euclidean}}$, normalised and mapped to range $[0,1]$:

\begin{equation}
    \mathcal{S}_{ssim}(v,\hat{v})=\exp\Bigl(-k \cdot \frac{\operatorname{DTW_{euclidean}}(S^\prime_{gen}, S^\prime_{ref})}{\max(T,\hat{T})}\Bigr),
\end{equation}

\noindent where $k$ was set to $k=5$, making the final metric $\mathcal{S}_{ssim}$ stricter. For the $\operatorname{DTW_{euclidean}}$ calculation, the fastDTW \cite{salvador_toward_2007} library is used for calculation efficiency, and the $\operatorname{euclidean}$ distance is used to calculate the distance of the elements in the two sequences.

Then to get $\mathcal{S}_{sem}$, each of the two RGB frame sets was encoded with CLIP embeddings and $\ell_2$ normalised to get semantic encoding for generated video $\mathcal{E}$ and reference video $\mathcal{\hat{E}}$:

\begin{equation}
    \mathcal{E} = \frac{f_{\text{CLIP}}(I)}{||f_{\text{CLIP}}(I)||_2}, \quad
    \mathcal{\hat{E}} = \frac{f_{\text{CLIP}}(\hat{I})}{||f_{\text{CLIP}}(\hat{I})||_2} ,
\end{equation}

\noindent where $f_{\text{CLIP}}$ represents the CLIP encoder from the ViT-L/14 \cite{radford_learning_2021} model. Then, $\operatorname{DTW}$ is calculated comparing $\mathcal{E}$ and $\mathcal{\hat{E}}$, normalised and mapped to range $[0,1]$ to get $\mathcal{S}_{sem}$ using a custom distance calculation $d(e,\hat{e})$:

\begin{equation}
    \mathcal{S}_{sem}(v,\hat{v})=\exp\Bigl(-\frac{\operatorname{DTW}_{d(e,\hat{e})}(\mathcal{E}, \mathcal{\hat{E}})}{\max(T,\hat{T})}\Bigr) ,
\end{equation}

\noindent where
$
d(e,\hat{e})=1-\langle e, \hat{e} \rangle \text{ and  } (e \in \mathcal{E}, \hat{e} \in \mathcal{\hat{E}}).
$
Then, the final visual reward $\mathcal{R_V}$ is obtained calculating the geometric mean of $\mathcal{S}_{ssim}$ and $\mathcal{S}_{sem}$:

\begin{equation}
    \mathcal{R_V}(v,\hat{v}) = \sqrt{\mathcal{S}_{ssim}(v,\hat{v})  \cdot \mathcal{S}_{sem}(v,\hat{v})} .
\end{equation}

\subsubsection{\texorpdfstring{Unified Reward ($\mathcal{R}$)}{R}}
Unifying both text reward $\mathcal{R_T}$ and visual reward $\mathcal{R_V}$ unified reward $\mathcal{R}$ is obtained as below:
\begin{equation}
    \mathcal{R}(y,v,\hat{y},\hat{v})=\lambda_\mathcal{T} \mathcal{R_T}(y,\hat{y})+\lambda_\mathcal{V} \mathcal{R_V}(v,\hat{v}) ,
\end{equation}

and from Equations \ref{eq:code-extraction} and \ref{eq:manim-renderer},

\begin{equation}
    \mathcal{R}(x,c,\hat{y})=\lambda_\mathcal{T} \mathcal{R_T} \left[\mathcal{H}(c),\hat{y}\right]+\lambda_\mathcal{V} \mathcal{R_V}\left[\mathcal{M}\left(\mathcal{H}(c)\right),\mathcal{M}(\hat{y})\right] ,
\end{equation}

\noindent where $\lambda_\mathcal{T}$ and $\lambda_\mathcal{V}$ are later set to $\lambda_\mathcal{T}=0.2$ and $\lambda_\mathcal{V}=0.8$. The goal of the RL training cycle is to enable the model to ground its knowledge in the visual results. However, the visual result was produced only when the code generated valid Manim code, resulting in a reward signal with high variance. A portion of the Text Similarity had to be introduced to mitigate this issue, since $\mathcal{R_T}$ produces a signal for both valid and invalid Manim code. The $\lambda_\mathcal{T}$ and $\lambda_\mathcal{V}$ values are empirically set after evaluating the geometric and arithmetic means of $\mathcal{R_T}$ and $\mathcal{R_V}$, with $\lambda_\mathcal{T} \in \{0.0, 0.2, 0.5\}$, and the $(0.2,0.8)$ pair yielded the most stable results.

\subsection{GRPO-based RL Optimisation}
\label{subsec:grpo-optimisation}

Using the reward function $\mathcal{R}(x,c,\hat{y})$ formulated in section \ref{subsec:reward-function}, the final loss optimisation based on Dr GRPO \cite{liu_understanding_2025} can be summarised as below:

\begin{equation}
    \mathcal{L}_{\text{Dr. GRPO}}(\theta) = - \frac{1}{LG} \sum_{i=1}^G \sum_{t=1}^{|c_i|} l_{i,t},
\end{equation}

\noindent where $\mathcal{L}_{\text{Dr. GRPO}}(\theta)$ represents optimisation loss calculated at $i^{\text{th}}$ sample's $t^\text{th}$ token,

\begin{equation}
\label{eq:loss-per-token}
\begin{aligned}
    l_{i,t}
    &= \min\left[r_{i,t}(\theta) \, \hat{A}_{i,t},
               \operatorname{clip}(r_{i,t}(\theta), 1-\epsilon, 1+\epsilon) \, \hat{A}_{i,t}\right] \\[2pt]
    &\quad- \beta \, \mathbb{D}_{\mathrm{KL}}\bigl[\pi_\theta \Vert \pi_{\mathrm{ref}}\bigr]
\end{aligned}
,
\end{equation}

\noindent where

\begin{equation}
    r_{i,t}(\theta) =\frac{\pi_\theta(c_{i,t} \mid x, c_{i,< t})}{\pi_{\theta_{\text{old}}}(c_{i,t} \mid x, c_{i,< t})},
\end{equation}

\noindent and

\begin{equation}
    \label{eq:advantage-calculation}
    \hat{A}_{i,t} = \frac{\mathcal{R}(x_i,c_i,\hat{y}_i) - \text{mean}[\mathcal{R}(x,c,\hat{y})]}{\text{std}[\mathcal{R}(x,c,\hat{y})]} .
\end{equation}

The mean and standard deviation (std) in Equation \ref{eq:advantage-calculation} are calculated over a sample of 8 produced for each $x_i$, setting $G=8$ and maximum completion length $L$, is set $L\approx2048\times0.8$. Further, $\beta=0.005$, and $\epsilon=0.2$ in Equation \ref{eq:loss-per-token}. $G$ is selected based on the available resources during training and the stability of the GRPO training cycle, and $L$ is determined after analysing the expected token lengths of completed examples. $\beta$ and $\epsilon$ are selected based on recommended values \cite{liu_understanding_2025}. GRPOTrainer from TRL is used for the above optimisations in the code implementation.

\subsection{Inference-time enhancements}
\begin{figure}[t]
    \centering
    \includegraphics[width=1\linewidth]{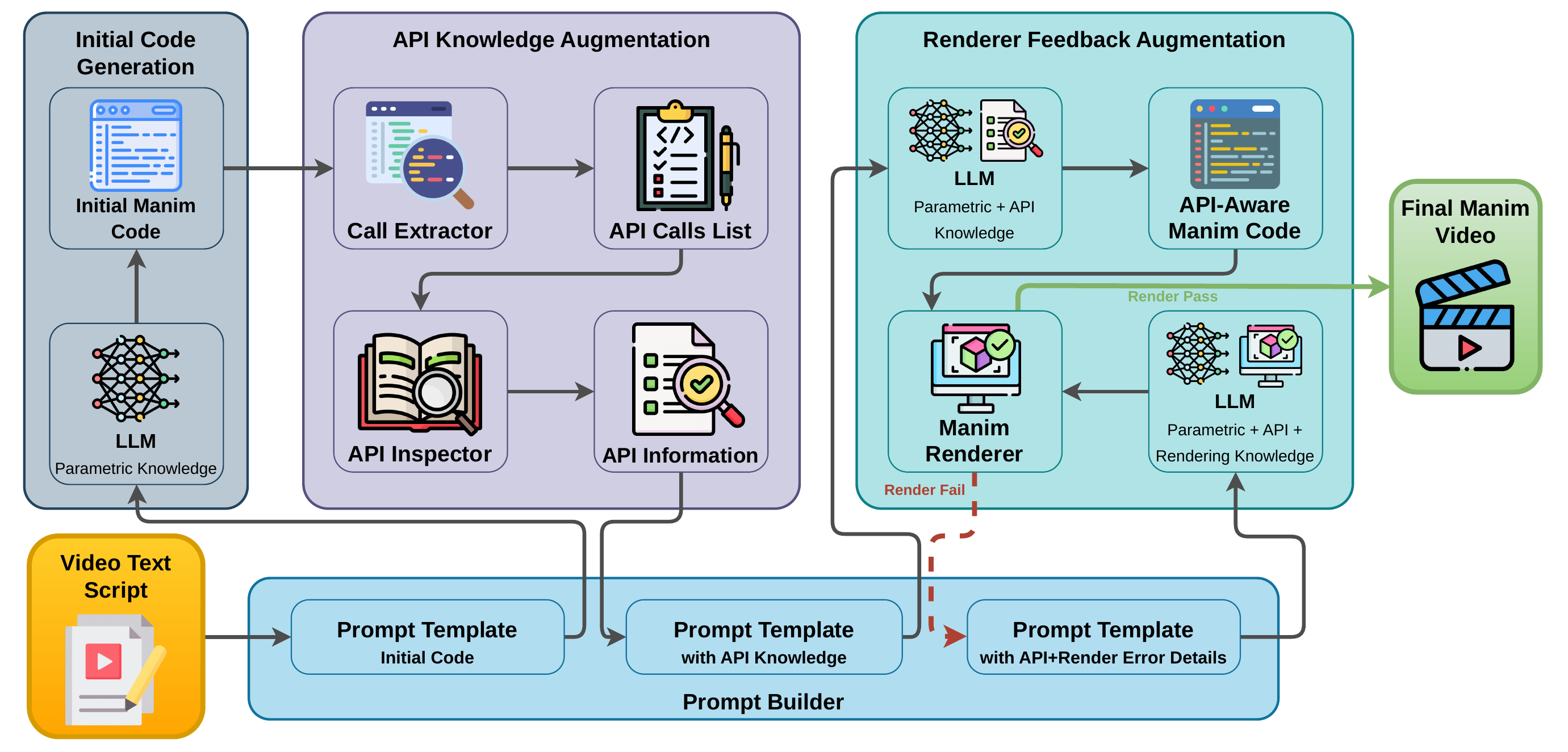}
    \caption{Inference Pipeline of the ManimAgent.}
    \label{fig:inference-pipeline}
\end{figure}

To investigate the influence of inference-time enhancements, two major inference-time enhancements were introduced in this study: Renderer-in-the-loop (RITL) and RITL with API Documentation (RITL-DOC). As shown in Figure~\ref{fig:inference-pipeline}, ManimAgent uses these techniques to iteratively validate and fix, involving the renderer within the loop, while embedding relevant API Documentation based on the initial Manim code generated by the LLM.

\subsubsection{Renderer in the loop (RITL)}
\label{subsubsec:ritl}
Let $d$ represent the textual description of the animation, $\mathcal{M}$ the LLM, and $\operatorname{ManimRender}(\cdot)$ the Python-based Manim Community Edition (ManimCE) 0.19.0 rendering engine that returns a status $s$, where $s\in\{\texttt{success},\texttt{fail}\}$ together with a renderer error output $e$. 

Given initial code $c_1\leftarrow\mathcal{M}(d)$, RITL iteratively loops for a maximum of $K$ rounds, transforming $(d, c_k, e_k)$ into an improved code $c_{k+1}$ and re-rendering it using $\operatorname{BuildPrompt_{RITL}}(\cdot)$. As shown in Algorithm \ref{alg:ritl-doc}, the loop terminates either upon finding a successful render $s=\texttt{success}$ or when reaching $k=K$. To reduce noise in the formulated prompt $p_{\mathrm{RITL}}$, the last ten lines of the render error log are used, which are sufficient to highlight the actual issue in the code.

\subsubsection{RITL-DOC - RITL with retrieved API Documentation}
\label{subsubsec:ritl-doc}
As shown in the Algorithm \ref{alg:ritl-doc}, RITL-DOC follows a functionally similar approach to RITL in Section~\ref{subsubsec:ritl}, but additionally incorporates API documentation $\mathcal{R}_k$ retrieved via the extracted API calls set $\mathcal{A}_k=\{a_{k,1},\dotsc, a_{k,m}\}$ when $m$ number of API calls are extracted using $\operatorname{BuildPrompt_{RITL-DOC}}$.

\begin{algorithm}[t]
\caption{Feedback + RAG-Augmented Iterative Code Generation (RITL/RITL-DOC)}
\label{alg:ritl-doc}
\begin{algorithmic}[1]
\Require Textual description $d$, maximum RITL rounds $K$, model $\mathcal{M}$, Manim API knowledge base $\mathcal{D}$
\Ensure Generated Manim code $c$, render status $s$
\State $c_1 \leftarrow \mathcal{M}(d)$ \Comment{Initial code generation}
\State $s_1, e_1 \leftarrow \operatorname{ManimRender}(c_1)$ \Comment{Render and capture status/errors}
\For{$k = 1$ \textbf{to} $K$}
    \If{$s_k = \texttt{success}$}
        \State \textbf{break} \Comment{Rendering succeeded; stop}
    \EndIf
    \If{$\operatorname{RITL-DOC}$}
        \State $\mathcal{A}_k \leftarrow \operatorname{ExtractAPICalls}(c_k)$ \Comment{Identify Manim API calls in code}
        \State $ \mathcal{R}_k \leftarrow \operatorname{RetrieveDocs}(\mathcal{A}_k,\ \mathcal{D}) $ \Comment{Relevant API documentation}
        \State $ p \leftarrow \operatorname{BuildPrompt_{RITL-DOC}}(d,\ c_k,\ e_k,\ \mathcal{R}_k) $ \Comment{Prompt with renderer output \& API documentation}
    \Else
        \State $p \leftarrow \operatorname{BuildPrompt_{RITL}}(d,\ c_k,\ e_k)$ \Comment{Prompt with renderer output}
    \EndIf
    \State $c_{k+1} \leftarrow \mathcal{M}(p)$ \Comment{Regenerate code}
    \State $s_{k+1}, e_{k+1} \leftarrow \operatorname{ManimRender}(c_{k+1})$
\EndFor
\State \Return $c,\; s$
\end{algorithmic}
\end{algorithm}

The API calls were extracted from the function header comments in Manim's Python library files. To address the prompt $p_{\mathrm{RITL-DOC}}$ being excessively large, only the parameter details of the API functions were included, while examples usually found in the function comments were left out. Using this approach, RITL-DOC bypasses the need for an LLM-based or a vector-based retriever by directly extracting API information from the initial code.

\subsection{Experimental Setup}
The following sections present the technical details of the experimental setup built for this study.

\subsubsection{Training and Evaluation Setup}
\label{subsubsec:training-setup}
The ManimBench \cite{silva_large_2026} dataset was used in this study to investigate the research question, as it includes human-written Manim code snippets alongside natural-language descriptions of the resulting videos. The dataset covers all Manim APIs via code examples, providing a comprehensive knowledge base for learning. Of the total 417 code samples, the official test set of 100 samples was kept for evaluation, while the remaining samples were used for training. The training prompt was modelled on the ManimBench study and employed ``System'' and ``User'' roles where available in the model tokens.

The models used for the fine-tuning experiments were selected from the Hugging Face Unsloth repository \cite{daniel_han_unsloth_2023}, mainly because it supports resource-efficient fine-tuning techniques. Seventeen models with fewer than 30B parameters were chosen for the experiments from various model families: Qwen 3 \cite{yang_qwen3_2025}, Qwen 2.5 \cite{hui_qwen25-coder_2024}, SeedCoder \cite{seed_seed-coder_2025}, LLaMA 3.1 \cite{grattafiori_llama_2024}, Mistral 3.2, and Ministral 3 \cite{liu_ministral_2026}. In selecting these models, performance on code generation tasks in BigCodeBench \cite{zhuo_bigcodebench_2024}, and their availability in the Unsloth model collection were considered. The following LoRA parameters were used in the training pipeline: Rank $r = 8$; scaling $\alpha = r\times2$; and targeting the Query, Key, Value and Output Projection modules.

For both SFT and GRPO training cycles, the batch size was set to 16. The LoRA adapters were trained with 16-bit precision, using the 8-bit AdamW optimiser while keeping the base model frozen at 4-bit Unsloth quantisation. A rank of 8 was selected to balance the adapter capacity with the dataset size. During the SFT phase, the models were trained for 2 epochs with a learning rate of $2e-6$ and a linear learning-rate reduction schedule. A weight decay of 0.01 and a warmup ratio of 0.1 were set to prevent overfitting. The GRPO training cycle was run with a reduced learning rate of $5e-7$. GRPO training is performed on top of the SFT model to mitigate reward hacking.

All training and evaluation experiments were conducted on an Ubuntu 24.04 LTS server with 64GB of RAM, an Nvidia RTX 5090 32GB VRAM, and 512GB of SSD storage. ManimCE 0.19.0 and Python 3.12.12 were used for rendering and data analysis. The code used for the experiments is available as open-source on GitHub \footnote{The code is available in GitHub: \url{https://github.com/SuienS/manim-trainer}}.

\subsubsection{Evaluation Metrics}
This study employed various evaluation metrics to assess both the generated code and the resulting video from functional and semantic perspectives. For code evaluation, CodeBLEU \cite{ren_codebleu_2020}, CodeBERT Similarity \cite{feng_codebert_2020}, N-gram Match, Syntax Match, and Abstract Syntax Tree (AST) Distance were utilised. Using these metrics, the generated code is compared with the expected reference code in the dataset. For video evaluation, Structural Similarity (SSIM) \cite{ndajah_ssim_2010} and the Visual Semantic Similarity using CLIP embeddings, as described in Section~\ref{subsubsec:visual-reward-calc}, were employed. Similarly, in the code evaluation, the resulting video from the generated code is compared with the resulting video of the expected reference code. 

However, to analyse the results more effectively, the above-mentioned metrics were combined into three key metrics: Visual Similarity (VS), CodeBERTBLEU (CBB), and Manim Render Success Rate (RSR). The VS and CBB are computed identically to $\mathcal{R_V}$ and $\mathcal{R_T}$ as described in Section~\ref{subsec:reward-function}. Manim RSR represents the fraction of generated code that successfully produces a Manim video.

\section{Results and Discussion}
\label{sec:results}

\subsection{Overall effect of SFT and GRPO on performance under Vanilla Inferencing}
The study evaluated 17 open-source sub-30B models using three key metrics that collectively represent the LLMs' Manim Code Generation ability. The Table~\ref{tbl:vanilla-full-performance} shows that SFT and GRPO consistently improved the performance of LLMs in terms of the RSR and VS, demonstrating the benefit of both fine-tuning strategies. However, CBB remained relatively similar across all model families and sizes, except for the Ministral 3 family models. This behaviour suggests that the models learned to fix failing code rather than write new code under both SFT and GRPO with the given dataset. It is notable that, during vanilla inference, 8 models performed best with GRPO, whereas 9 models achieved their best at the SFT stage. However, the highest visual performance across all models was observed with the SeedCoder 8B model under GRPO with an RSR of 72\%. 

\begin{figure}[t]
    \centering
    \includegraphics[width=1\linewidth]{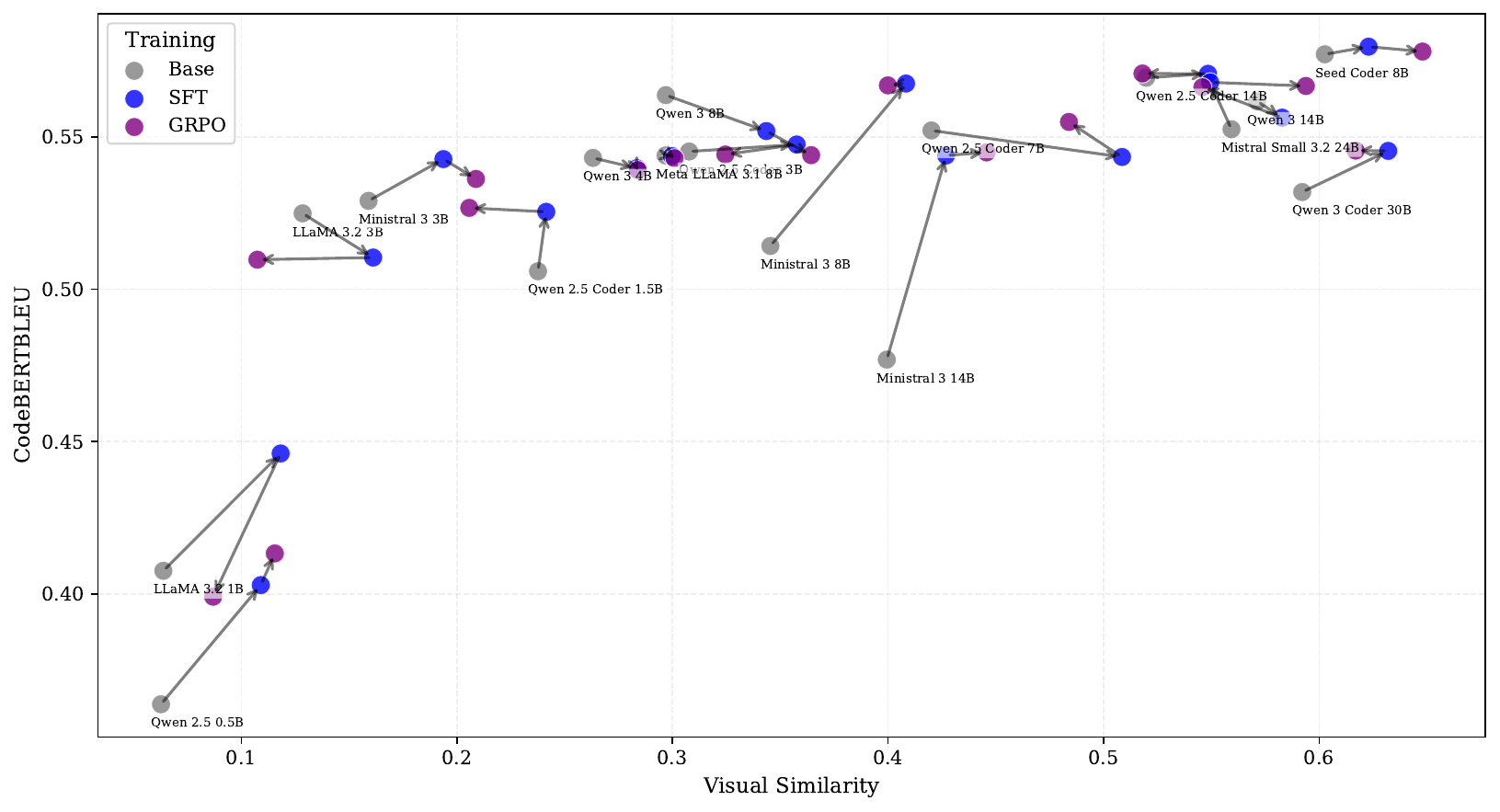}
    \caption{Behaviour of CodeBERTBLEU score against Visual Similarity during the training cycles.}
    \label{fig:visual_vs_code_scatter}
\end{figure}

The Figure~\ref{fig:visual_vs_code_scatter} illustrates that improvements in CBB typically occur during the SFT stage, whereas enhancements in VS typically occur during the GRPO training cycle. This aligns with DeepSeek-R1 training observations \cite{guo_deepseek-r1_2025}. However, the results also highlight the unpredictability of smaller models during training, where models such as LLaMA 3.2 1B improved performance on both metrics, whereas the Qwen 2.5 0.5B model consistently improved its performance across the two training cycles.

\begin{table}[t]
\centering
\caption{Evaluation summary of all the fine-tuned models under vanilla inferencing. The scores present the percentage visual and code similarity value under VS and CBB columns, and the percentage of generated samples successfully rendered under Render Success Rate.}
\label{tbl:vanilla-full-performance}
\begin{tiny}
\begin{tabular}{@{}lccccccccc@{}}
\toprule
\multirow{2}{*}{\textbf{Model}} &
  \multicolumn{3}{c}{\textbf{Visual Similarity}} &
  \multicolumn{3}{c}{\textbf{CodeBERTBLEU}} &
  \multicolumn{3}{c}{\textbf{Render Success Rate}} \\ \cmidrule(l){2-10} 
 &
  \textbf{Base} &
  \textbf{SFT} &
  \textbf{GRPO} &
  \textbf{Base} &
  \textbf{SFT} &
  \textbf{GRPO} &
  \textbf{Base} &
  \textbf{SFT} &
  \textbf{GRPO} \\ \midrule
Qwen 2.5 0.5B         & 6.3  & 10.9 & \textbf{11.6} & 36.4 & 40.3 & 41.3 & 7.0  & 12.0 & 13.0 \\
LLaMA 3.2 1B          & 6.4  & \textbf{11.8} & 8.7  & 40.8 & 44.6 & 39.9 & 7.0  & 13.0 & 10.0 \\
LLaMA 3.2 3B          & 12.8 & \textbf{16.1} & 10.7 & 52.5 & 51.0 & 51.0 & 14.0 & 18.0 & 12.0 \\
Ministral 3 3B        & 15.9 & 19.4 & \textbf{20.9} & 52.9 & 54.3 & 53.6 & 18.0 & 22.0 & 23.0 \\
Qwen 2.5 Coder 3B     & 30.8 & \textbf{35.8} & 32.4 & 54.5 & 54.8 & 54.4 & 34.0 & 39.0 & 36.0 \\
Qwen 3 4B             & 26.3 & 28.4 & \textbf{28.4} & 54.3 & 54.0 & 53.9 & 29.0 & 31.0 & 31.0 \\
Qwen 2.5 Coder 1.5B   & 23.8 & \textbf{24.1} & 20.6 & 50.6 & 52.5 & 52.7 & 27.0 & 27.0 & 23.0 \\
Qwen 2.5 Coder 7B     & 42.0 & \textbf{50.8} & 48.4 & 55.2 & 54.3 & 55.5 & 46.0 & 55.0 & 53.0 \\
Meta LLaMA 3.1 8B     & 29.7 & 30.0 & \textbf{30.1} & 54.4 & 54.4 & 54.3 & 33.0 & 33.0 & 33.0 \\
Ministral 3 8B        & 34.5 & \textbf{40.8} & 40.0 & 51.4 & 56.8 & 56.7 & 40.0 & 44.0 & 43.0 \\
Qwen 3 8B             & 29.7 & 34.4 & \textbf{36.4} & 56.4 & 55.2 & 54.4 & 33.0 & 38.0 & 40.0 \\
SeedCoder 8B         & 60.3 & 62.3 & \textbf{\underline{64.8}} & 57.7 & 58.0 & 57.8 & 67.0 & 69.0 & 72.0 \\
Ministral 3 14B       & 39.9 & 42.7 & \textbf{44.6} & 47.7 & 54.4 & 54.5 & 45.0 & 46.0 & 48.0 \\
Qwen 2.5 Coder 14B    & 52.0 & \textbf{54.8} & 51.8 & 56.9 & 57.1 & 57.1 & 57.0 & 60.0 & 57.0 \\
Qwen 3 14B            & 57.1 & \textbf{58.3} & 54.6 & 56.2 & 55.6 & 56.6 & 64.0 & 65.0 & 60.0 \\
Mistral Small 3.2 24B & 55.9 & 54.9 & \textbf{59.4} & 55.2 & 56.8 & 56.7 & 63.0 & 59.0 & 64.0 \\
Qwen 3 Coder 30B (A3B)& 59.2 & \textbf{63.2} & 61.7 & 53.2 & 54.5 & 54.6 & 66.0 & 70.0 & 68.0 \\
\midrule
Qwen 3 Coder Next 80B & 63.7 & -    & -    & 54.6 & -    & -    & 72.0 & -    & -    \\
OpenAI GPT-4.1        & 68.6 & -    & -    & 55.9 & -    & -    & 77.0 & -    & -    \\ 
\bottomrule
\end{tabular}
\end{tiny}
\end{table}

\subsection{Effect of Parameter Count on the Fine-tuning strategy}

\begin{figure}[t]
    \centering
    \includegraphics[width=1\linewidth]{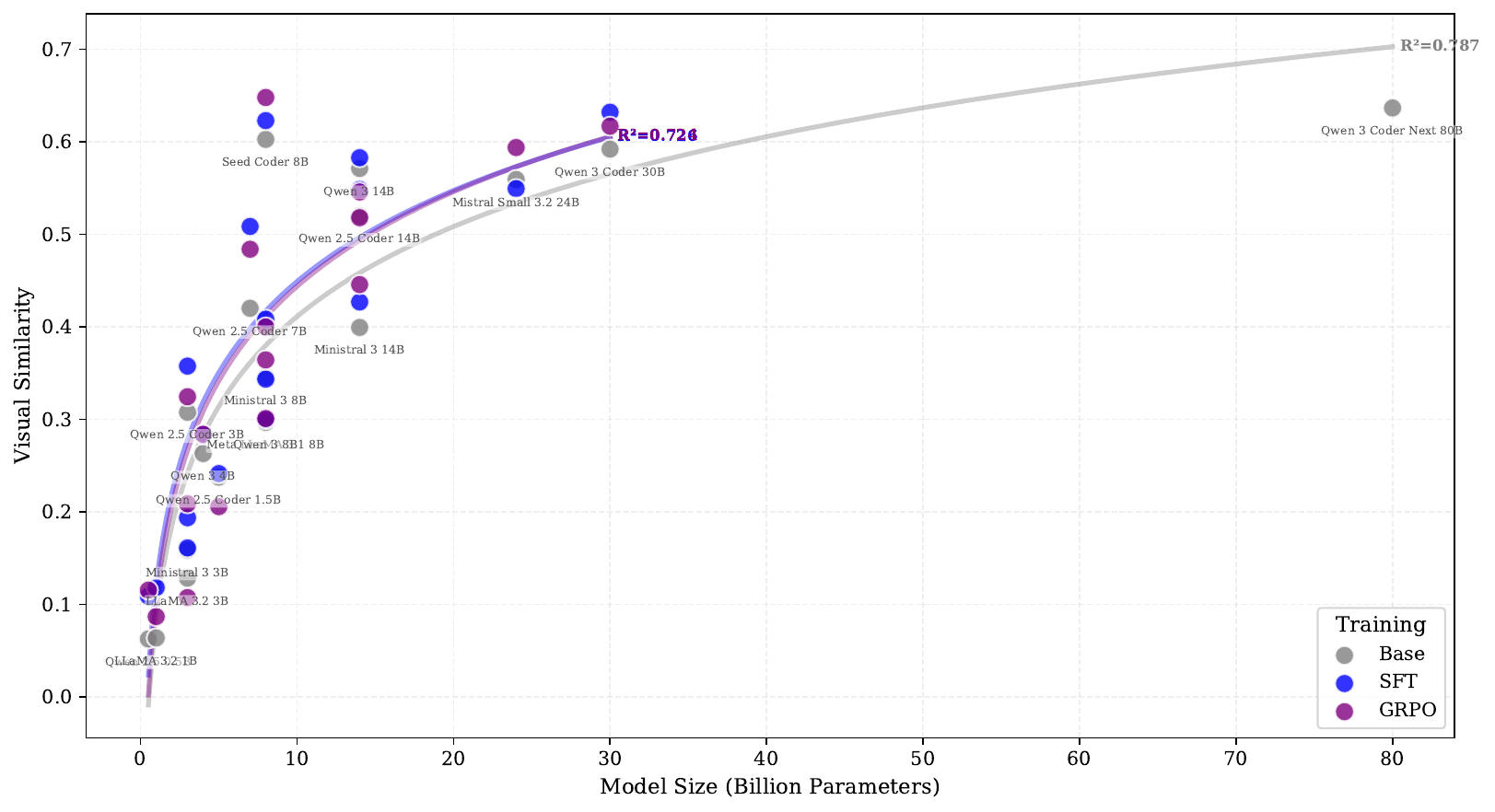}
    \caption{Scaling Behaviour of the Visual Similarity in the fine-tuned LLMs.}
    \label{fig:scaling-behaviour}
\end{figure}

The findings shown in Table~\ref{tbl:vanilla-full-performance} indicate that VS consistently improved with either SFT, GRPO, or both across all model sizes. However, the percentage improvement decreased as the parameter count increased. SeedCoder 8B performs the best in VS, as shown in Figure~\ref{fig:scaling-behaviour}, despite having only 8B parameters. Under GRPO, the SeedCoder 8B model has surpassed even the newer Qwen 3 Coder Next 80B MoE model and is comparable to GPT-4.1, highlighting its efficiency and reliability. Other models with a parameter count range of 7B-8B demonstrate significant improvements with SFT and GRPO.

For example, Qwen 2.5 Coder 7B, Ministral 8B, and Qwen 3 8B models demonstrated notable improvements with either SFT or GRPO, respectively increasing their Base VS from 42.0, 34.5, and 29.7 to 50.8, 40.8, and 36.4. However, the performance gains of $>$14B models were comparatively lower. This observation is consistent with the findings of \cite{hoffmann_training_2022}, which note that larger models require proportionally larger training datasets to improve. Across families, the 7B-8B range emerges as an efficiency sweet spot, achieving strong VS gains while maintaining comparable performances to larger models.

\begin{figure}[t]
    \centering
    \includegraphics[width=1\linewidth]{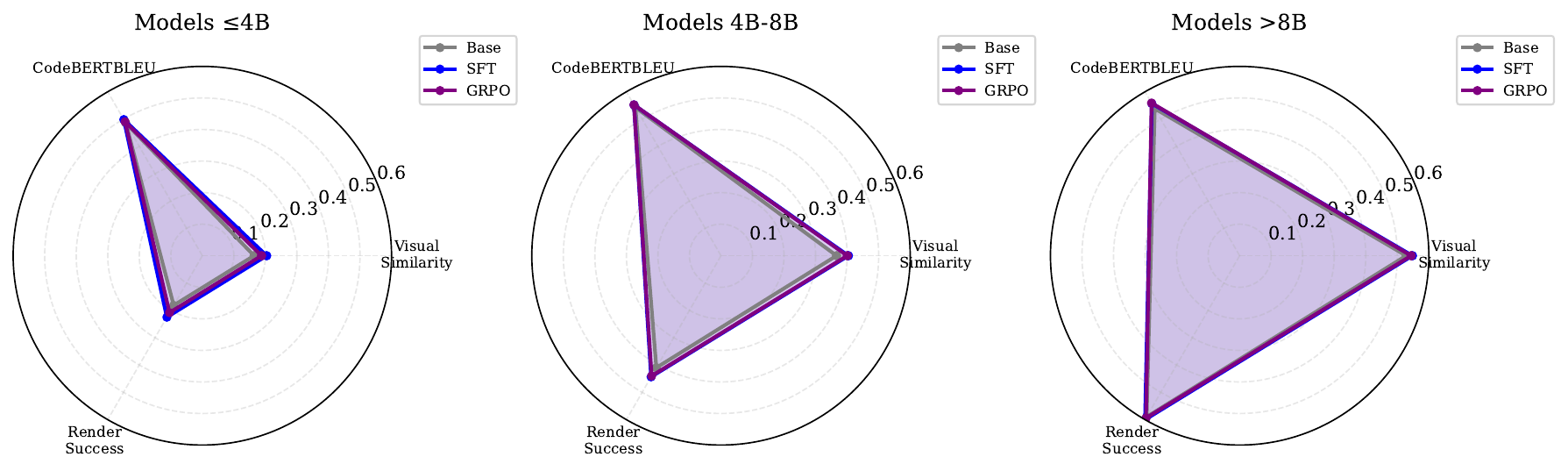}
    \caption{Radar plot of CodeBERTBLEU, Visual Similarity, and Render Success Rate of the models grouped by model size category. CBB remains relatively unchanged across groups, while VS and RSR regions expand.}
    \label{fig:radar-evals-size-grouped}
\end{figure}

Figure~\ref{fig:radar-evals-size-grouped} shows that, regardless of the fine-tuning strategy, all models of different sizes display similar performance in CBB, while their performance is more notable and improves significantly in VS and RSR with higher parameter counts. The figure also indicates that SFT slightly outperforms GRPO for models with 4B parameters or fewer, whereas GRPO outperforms SFT for larger models.

These trends suggest a capacity threshold for the RL training cycle. GRPO appears to benefit models that already possess sufficient task competence after SFT, whereas smaller models ($\le$4B) may struggle to generate informative samples during RL optimisation, thereby limiting GRPO's influence relative to SFT.

\subsection{Overall effect of RITL and RITL-DOC on the Model Performance}

Investigating the effect of inference-time strategies, RITL and RITL-DOC techniques were introduced as explained in Sections \ref{subsubsec:ritl} and \ref{subsubsec:ritl-doc} at inference time for all 17 models across their Base, SFT, and GRPO variants. As shown in Table~\ref{tbl:ritl-full-performance}, RITL consistently enhanced both VS and RSR for most models. Qwen 3 8B increased its baseline VS by more than 10 percentage points, while the significantly larger Qwen 3 Coder 30B model scored 82.2\% under SFT, up from 74.5\% VS under Base, making it the best performing model in the table, even surpassing the much larger Qwen 3 Coder Next 80B model and GPT-4.1.

\begin{table}[t]
\centering
\caption{Evaluation summary of all the fine-tuned models under RITL inference}
\label{tbl:ritl-full-performance}
\begin{tiny}
\begin{tabular}{@{}lccccccccc@{}}
\toprule
\multirow{2}{*}{\textbf{Model (with RITL)}} & \multicolumn{3}{c}{\textbf{Visual Similarity}} & \multicolumn{3}{c}{\textbf{CodeBERTBLEU}} & \multicolumn{3}{c}{\textbf{Render Success Rate}} \\ \cmidrule(l){2-10} 
                      & \textbf{Base} & \textbf{SFT}  & \textbf{GRPO} & \textbf{Base} & \textbf{SFT}  & \textbf{GRPO} & \textbf{Base} & \textbf{SFT}  & \textbf{GRPO} \\ \midrule
Qwen 2.5 0.5B         & 8.1  & 11.3 & \textbf{12.4} & 46.2 & 45.1 & 44.5 & 9.0  & 13.0 & 14.0 \\
LLaMA 3.2 1B          & 9.7  & 10.8 & \textbf{12.4} & 49.3 & 47.6 & 47.7 & 11.0 & 12.0 & 14.0 \\
LLaMA 3.2 3B          & 14.3 & 15.8 & \textbf{17.2} & 53.2 & 53.6 & 53.5 & 16.0 & 18.0 & 19.0 \\
Ministral 3 3B        & 21.2 & 19.4 & \textbf{24.4} & 28.4 & 53.4 & 55.2 & 24.0 & 22.0 & 27.0 \\
Qwen 2.5 Coder 3B     & 34.0 & \textbf{36.8} & 35.5 & 54.7 & 53.9 & 54.8 & 38.0 & 40.0 & 39.0 \\
Qwen 3 4B             & 33.5 & 35.8 & 29.8 & 54.6 & 54.8 & 53.7 & 37.0 & 40.0 & 33.0 \\
Qwen 2.5 Coder 1.5B   & 24.0 & 22.8 & \textbf{25.2} & 51.4 & 51.3 & 52.2 & 27.0 & 26.0 & 29.0 \\
Qwen 2.5 Coder 7B     & \textbf{61.0} & 58.0 & 58.8 & 54.6 & 55.6 & 54.9 & 66.0 & 63.0 & 64.0 \\
Meta LLaMA 3.1 8B     & 34.1 & 31.3 & \textbf{39.8} & 54.5 & 55.0 & 54.7 & 39.0 & 37.0 & 45.0 \\
Ministral 3 8B        & 45.9 & 47.3 & \textbf{48.0} & 52.0 & 56.9 & 56.1 & 53.0 & 51.0 & 52.0 \\
Qwen 3 8B             & 37.0 & 40.4 & \textbf{47.3} & 56.2 & 55.9 & 56.6 & 41.0 & 44.0 & 53.0 \\
SeedCoder 8B         & 66.4 & 68.4 & \textbf{69.2} & 57.5 & 57.9 & 57.8 & 74.0 & 76.0 & 77.0 \\
Ministral 3 14B       & 49.9 & 55.9 & \textbf{57.8} & 52.2 & 55.8 & 55.9 & 56.0 & 61.0 & 63.0 \\
Qwen 2.5 Coder 14B    & 65.2 & \textbf{69.0} & 66.1 & 57.0 & 57.1 & 56.8 & 72.0 & 77.0 & 72.0 \\
Qwen 3 14B            & 67.8 & 62.7 & \textbf{64.0} & 56.3 & 55.8 & 55.4 & 76.0 & 69.0 & 71.0 \\
Mistral Small 3.2 24B & 63.1 & 60.4 & \textbf{63.4} & 55.1 & 57.2 & 56.4 & 71.0 & 65.0 & 68.0 \\
Qwen 3 Coder 30B (A3B)& 74.5 & \textbf{\underline{82.2}} & 80.6 & 53.3 & 54.8 & 54.1 & 83.0 & 90.0 & 89.0 \\
\midrule
Qwen 3 Coder Next 80B & 72.6 & -    & -    & 53.9 & -    & -    & 82.0 & -    & -    \\ 
OpenAI GPT-4.1        & 76.9 & -    & -    & 56.2 & -    & -    & 86.0 & -    & -    \\ 
\bottomrule
\end{tabular}
\end{tiny}
\end{table}

Table~\ref{tbl:ritl-doc-full-performance} presents the performance evaluations under RITL-DOC, which integrates the RITL loop with relevant API documentation. Under RITL-DOC, models with more than 7B parameters generally demonstrate further improvements over the baseline models, whereas smaller models tend to perform better with SFT than with GRPO. Qwen 3 8B and Mistral Small 3.2 24B models experienced a performance reduction during SFT and GRPO training cycles under RITL-DOC, although their performance remained higher than under RITL. For smaller models with less than 4B parameters, RITL occasionally outperformed RITL-DOC. However, Qwen 3 Coder 30B performed best overall across all models and strategies, with GRPO under RITL-DOC, achieving approximately 86\% of VS and 94\% of RSR after three RITL attempts. Notably, the RITL-DOC Qwen 3 Coder 30B model, with only 3B active parameters out of 30B total, surpassed OpenAI's GPT-4.1.

\begin{table}[t]
\centering
\caption{Evaluation summary of all the fine-tuned models under RITL-DOC inference}
\label{tbl:ritl-doc-full-performance}
\begin{tiny}
\begin{tabular}{@{}lccccccccc@{}}
\toprule
\multirow{2}{*}{\textbf{Model (RITL-DOC)}} & \multicolumn{3}{c}{\textbf{Visual Similarity}} & \multicolumn{3}{c}{\textbf{CodeBERTBLEU}} & \multicolumn{3}{c}{\textbf{Render Success Rate}} \\ \cmidrule(l){2-10} 
                      & \textbf{Base} & \textbf{SFT}  & \textbf{GRPO} & \textbf{Base} & \textbf{SFT}  & \textbf{GRPO} & \textbf{Base} & \textbf{SFT}  & \textbf{GRPO} \\ \midrule
Qwen 2.5 0.5B         & 8.8  & 7.2  & \textbf{8.8}  & 49.5 & 48.4 & 50.0 & 10.0 & 8.0  & 10.0 \\
LLaMA 3.2 1B          & 10.0 & 9.2  & \textbf{10.0} & 48.7 & 47.7 & 45.8 & 11.0 & 10.0 & 11.0 \\
LLaMA 3.2 3B          & 14.3 & 12.6 & \textbf{20.5} & 51.5 & 52.3 & 51.2 & 16.0 & 14.0 & 23.0 \\
Ministral 3 3B        & 20.7 & \textbf{31.4} & 30.5 & 47.6 & 56.0 & 55.4 & 24.0 & 35.0 & 34.0 \\
Qwen 2.5 Coder 3B     & 36.2 & \textbf{40.1} & 30.6 & 54.4 & 54.5 & 54.7 & 40.0 & 44.0 & 34.0 \\
Qwen 3 4B             & 25.2 & \textbf{31.9} & 30.2 & 55.0 & 54.3 & 54.1 & 28.0 & 35.0 & 33.0 \\
Qwen 2.5 Coder 1.5B   & 23.0 & \textbf{24.7} & 20.4 & 51.6 & 50.8 & 51.7 & 26.0 & 28.0 & 23.0 \\
Qwen 2.5 Coder 7B     & 60.6 & 63.5 & \textbf{64.9} & 54.8 & 55.0 & 55.1 & 66.0 & 69.0 & 71.0 \\
Meta LLaMA 3.1 8B     & 38.9 & 40.2 & \textbf{41.2} & 54.5 & 54.3 & 54.3 & 44.0 & 45.0 & 46.0 \\
Ministral 3 8B        & 50.2 & \textbf{55.4} & 54.6 & 52.3 & 57.2 & 56.9 & 58.0 & 60.0 & 59.0 \\
Qwen 3 8B             & \textbf{51.4} & 41.5 & 47.4 & 56.5 & 55.4 & 55.7 & 58.0 & 47.0 & 52.0 \\
SeedCoder 8B         & 65.6 & 69.2 & \textbf{69.2} & 57.3 & 57.5 & 57.4 & 73.0 & 77.0 & 77.0 \\
Ministral 3 14B       & 56.8 & \textbf{59.7} & 59.6 & 52.5 & 55.9 & 56.2 & 64.0 & 65.0 & 65.0 \\
Qwen 2.5 Coder 14B    & 66.0 & 66.4 & \textbf{67.1} & 56.6 & 56.9 & 56.9 & 73.0 & 73.0 & 74.0 \\
Qwen 3 14B            & 63.6 & 68.7 & \textbf{71.0} & 56.5 & 55.7 & 55.9 & 71.0 & 76.0 & 79.0 \\
Mistral Small 3.2 24B & \textbf{68.4} & 63.2 & 64.5 & 55.3 & 57.3 & 56.6 & 77.0 & 69.0 & 70.0 \\
Qwen 3 Coder 30B (A3B)& 77.7 & 80.8 & \textbf{83.0} & 53.5 & 54.4 & 53.7 & 87.0 & 89.0 & 92.0 \\
Qwen 3 Coder 30B (A3B) (@3) & 79.2 & 82.6 & \textbf{\underline{85.7}} & 53.1 & 53.7 & 54.2 & 89.0 & 92.0 & 94.0 \\
\midrule
Qwen 3 Coder Next 80B & 73.9 & -    & -    & 54.1 & -    & -    & 85.0 & -    & -    \\ 
OpenAI GPT-4.1        & 81.9 & -    & -    & 56.6 & -    & -    & 92.0 & -    & -    \\ 
OpenAI GPT-4.1  (@3)  & 82.8 & -    & -    & 56.1 & -    & -    & 94.0 & -    & -    \\ 
\bottomrule
\end{tabular}
\end{tiny}
\end{table}

The mean values in the Table~\ref{tbl:visual-sim-full-comp} suggest a monotonic progression: Vanilla Base, SFT, and GRPO to RITL Base, SFT, and GRPO to RITL-DOC Base, SFT, and GRPO. The results show that, overall, inference-level strategies deliver greater performance gains than fine-tuning alone. Additionally, RITL-DOC with GRPO emerges as the most effective combination of inference and training strategies throughout the table.

\begin{table}[t]
\centering
\caption{Comparison of the mean Visual Similarity score of all fine-tuned models across all combinations of training and inference strategy}
\label{tbl:visual-sim-full-comp}
\begin{tiny}
\begin{tabular}{@{}lccccccccc@{}}
\toprule
\textbf{Model} &
  \textbf{Base} &
  \textbf{SFT} &
  \textbf{GRPO} &
  \textbf{\begin{tabular}[c]{@{}c@{}}Base\\ RITL\end{tabular}} &
  \textbf{\begin{tabular}[c]{@{}c@{}}SFT\\ RITL\end{tabular}} &
  \textbf{\begin{tabular}[c]{@{}c@{}}GRPO\\ RITL\end{tabular}} &
  \textbf{\begin{tabular}[c]{@{}c@{}}Base\\ RITL-D.\end{tabular}} &
  \textbf{\begin{tabular}[c]{@{}c@{}}SFT\\ RITL-D.\end{tabular}} &
  \textbf{\begin{tabular}[c]{@{}c@{}}GRPO\\ RITL-D.\end{tabular}} \\ 
\midrule
{[}Mean{]}            & 34.3 & 37.6 & 36.8 & 41.7 & 42.8 & 44.2 & 43.4 & 45.0 & \textbf{\underline{45.5}} \\ \bottomrule
\end{tabular}
\end{tiny}
\end{table}

\subsubsection{Interaction between Training and Inference Strategies}
\label{subsubsec:interaction-train-vs-inference}

To distinguish the individual and combined effects of the two training and inference strategies, ablation experiments were conducted, grouped by the two strategies.

\begin{figure}[t]
    \centering
    \includegraphics[width=1\linewidth]{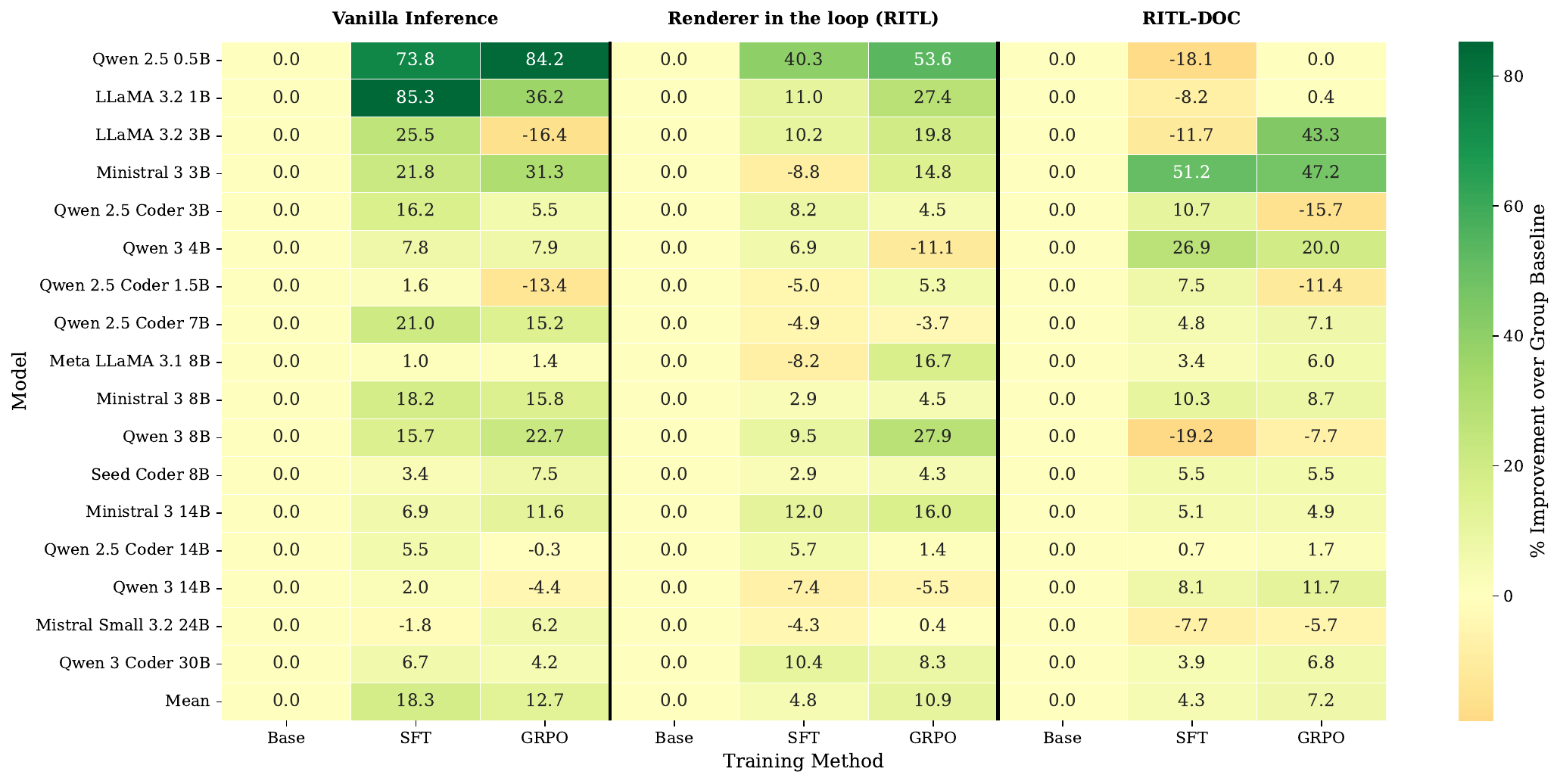}
    \caption{Ablation experiment results conducted to investigate the effect of training strategy on the visual similarity performance of the fine-tuned LLMs. The percentage improvement is presented, grouped by inference strategy, relative to the base model in each group.}
    \label{fig:inference-grouped-ablation}
\end{figure}

Figure~\ref{fig:inference-grouped-ablation} illustrates the effects of different training strategies categorised by inference approach. The figure indicates that the impact of the training strategy is particularly significant for smaller models across all inference methods. Based on the average percentage improvements, the influence of these strategies is more pronounced under the vanilla inference approach compared to RITL and RITL-DOC.

\begin{figure}[t]
    \centering
    \includegraphics[width=1\linewidth]{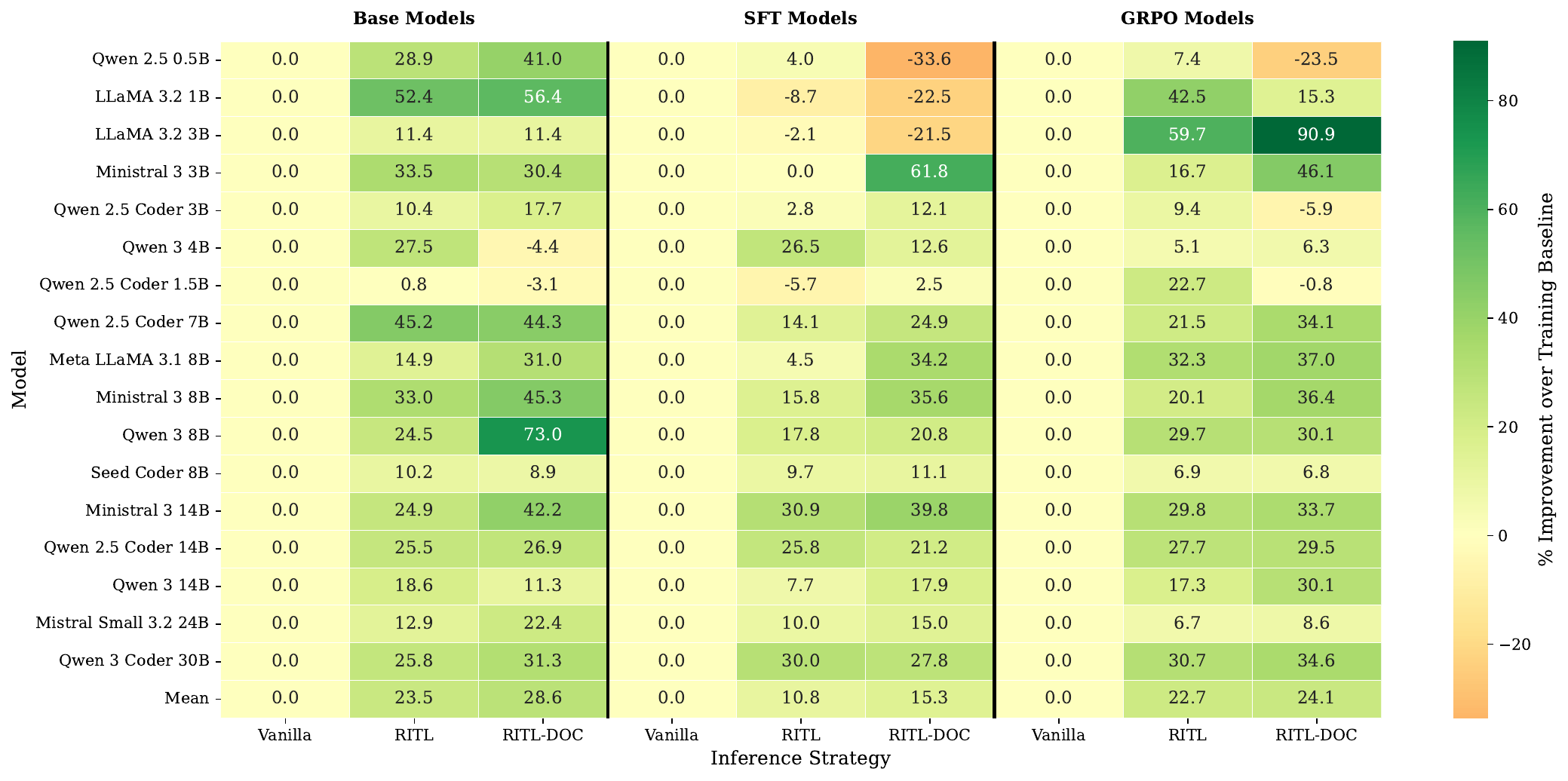}
    \caption{Ablation experiment results conducted to investigate the effect of inference strategy on visual similarity performance of the fine-tuned LLMs. The percentage improvement is presented, grouped by training strategy, relative to the vanilla inference strategy in each group.}
    \label{fig:training-grouped-ablation}
\end{figure}

Figure~\ref{fig:training-grouped-ablation} provides a complementary view to Figure~\ref{fig:inference-grouped-ablation}, demonstrating how the inference strategy affects different training methods. The figure shows that the RITL and RITL-DOC inference strategies have a stronger impact on models trained with Base and GRPO, but are less pronounced for SFT models compared to their respective baselines. Additionally, the figure indicates that smaller models decline in performance with RITL and RITL-DOC under SFT, whereas they perform better under RITL and RITL-DOC under GRPO, for example, the LLaMA 3.2 3B model. A likely explanation is context saturation since RITL-DOC can consume most of the available context length with dense API information overwhelming the smaller models.

\begin{figure}[t]
    \centering
    \includegraphics[width=0.8\linewidth]{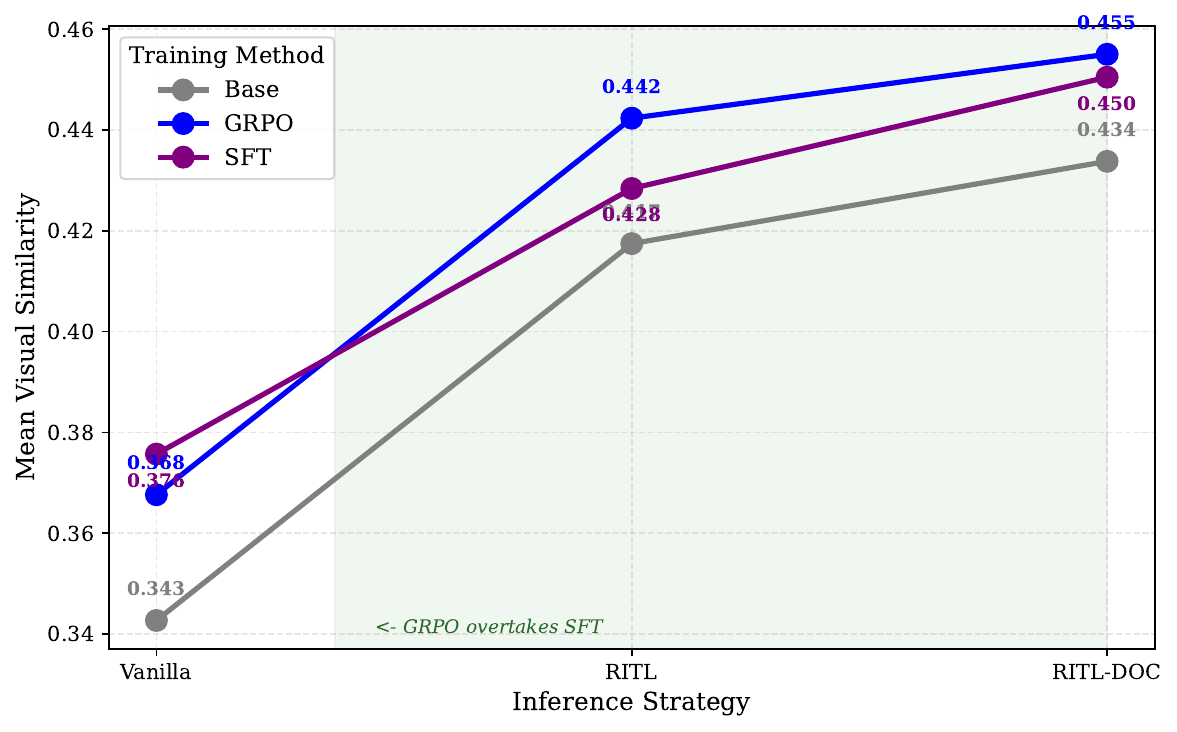}
    \caption{Variation of mean Visual Similarity of the fine-tuned models across the inference strategies.}
    \label{fig:mean-visual-vs-inf-strategy}
\end{figure}

Figure~\ref{fig:mean-visual-vs-inf-strategy} reveals a pattern where, overall, SFT slightly outperforms GRPO under vanilla inference, but GRPO improves and surpasses SFT under RITL and RITL-DOC. This pattern indicates that GRPO-trained models perform best with inference-time strategies such as RITL and RITL-DOC and are more receptive to iterative self-correction. This finding aligns with the nature of GRPO training, in which the model is trained to satisfy an external validation that closely matches the functioning of the feedback loop in RITL.

\subsubsection{Effect of Additional RITL Loops}
Throughout the experiments, only one iteration was used with RITL and RITL-DOC, giving the model a single opportunity to self-correct and focusing more on the model's ability to generate Manim code rather than on repairing faulty Manim code.

\begin{figure}
    \centering
    \includegraphics[width=1\linewidth]{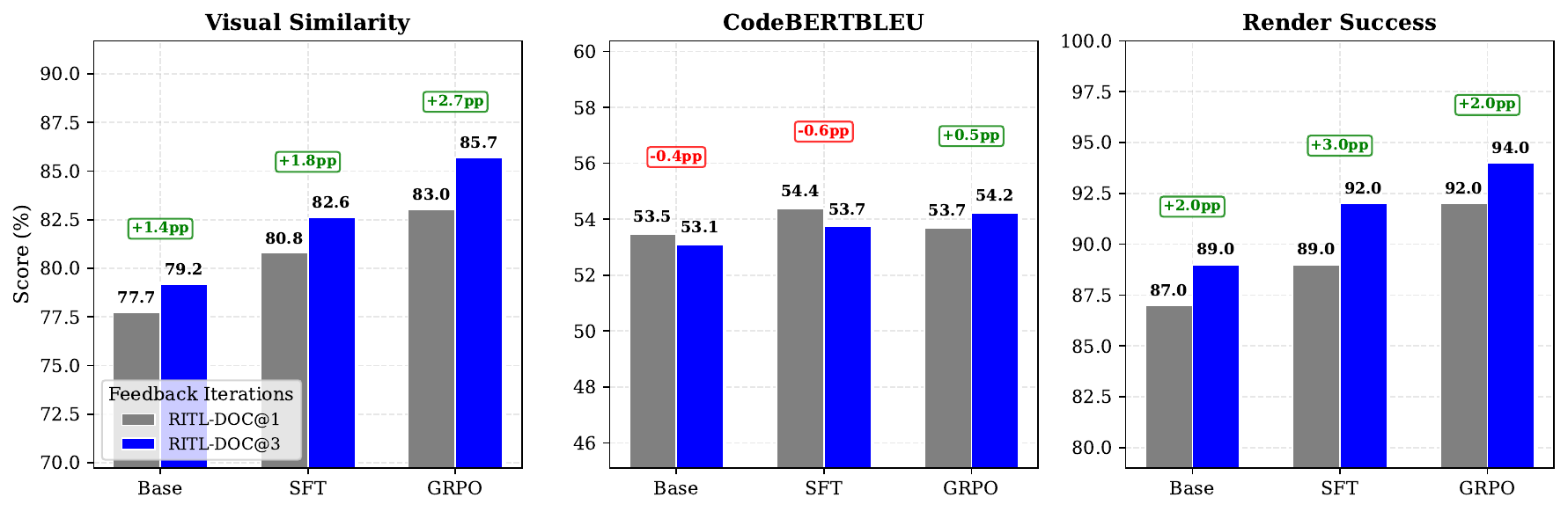}
    \caption{Comparison of performance between RITL-DOC@1 and RITL-DOC@3}
    \label{fig:fb1-fb3-comparison}
\end{figure}

However, to evaluate the benefits of additional self-correction attempts, the overall top-performing model, Qwen 3 Coder 30B, was tested under RITL-DOC with three correction attempts. As shown in Figure~\ref{fig:fb1-fb3-comparison}, the additional correction attempts consistently improved VS and RSR across all three training strategies, with greater gains seen with GRPO, confirming its role in self-correction ability, as also discussed in Section~\ref{subsubsec:interaction-train-vs-inference}.

However, Figure~\ref{fig:fb1-fb3-comparison} also shows that with Base and SFT model variants, the additional correction loops tend to lower the CodeBERTBLEU, indicating that they encourage the model to add fixes to the initial code, which may not be the most ideal way to generate that particular code for the provided prompt. GRPO models show a more balanced trend, with less CBB degradation under additional iterations.

\subsection{Correlation between Code and Visual Metrics}

\begin{figure}
    \centering
    \includegraphics[width=0.65\linewidth]{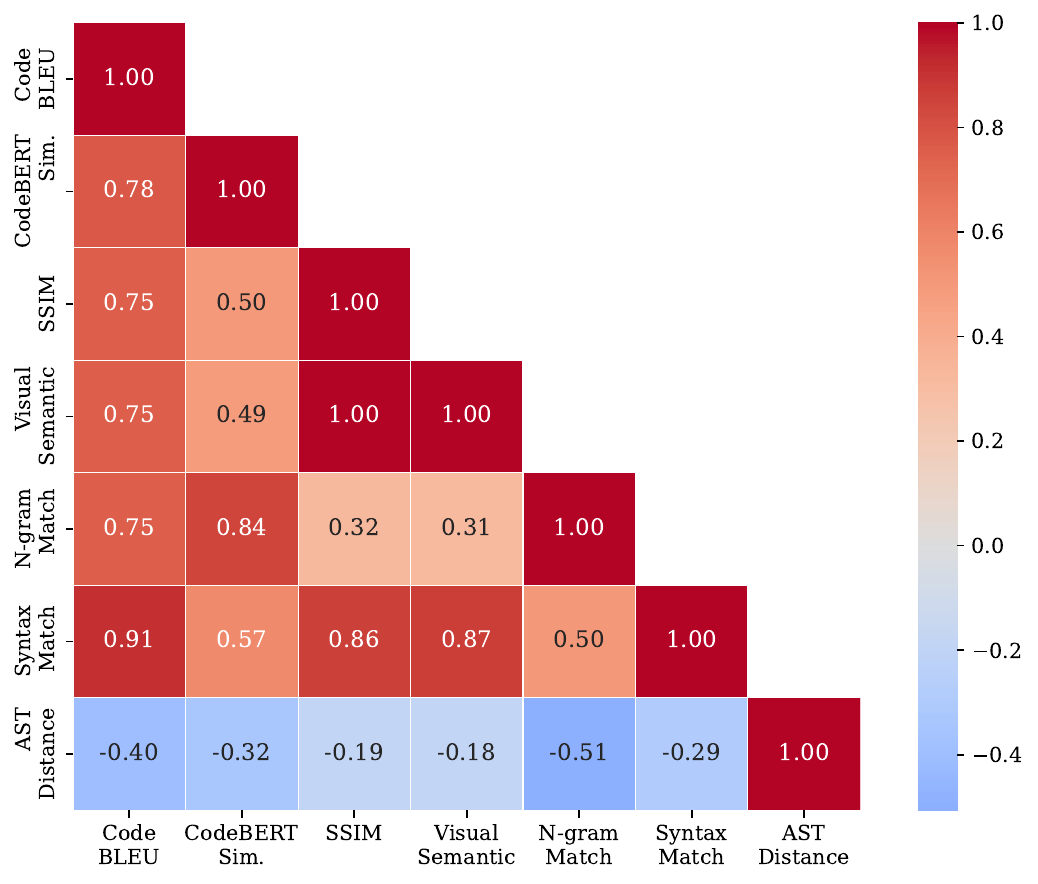}
    \caption{Spearman Rank Correlation Between Evaluation Metrics calculated for each model under all combinations of inference and training strategies.}
    \label{fig:spearman-corr-eval-metrics}
\end{figure}

Figure~\ref{fig:spearman-corr-eval-metrics} shows a Spearman rank correlation heatmap for the seven different evaluation metrics utilised in the previously mentioned experiments. Overall, within each metric category (code metrics and visual metrics), the correlation is stronger, as anticipated. For instance, although computed using two different algorithms, the Visual Semantic Score and the SSIM score are highly correlated. Similarly, CodeBLEU and CodeBERT Similarity exhibit high correlation. CodeBLEU exhibits moderate correlations with all other metrics, except AST Distance.

Nevertheless, the cross-family correlations between code and visual metrics are relatively weaker. This indicates that models may generate code that produces a video similar to the reference video while using structurally different code, or conversely, generate structurally similar code that ultimately fails to render. The figure also indicates no correlation of the AST distance with any other metric, implying that it is unreliable for the Manim code generation evaluation. One plausible reason could be Manim's object-oriented scene structure, which remains relatively consistent across all Manim scripts.

\subsubsection{Agreement of VS and CBB across Training and Inference Strategies}
To investigate the connection between VS and CBB, Table~\ref{tbl:result-correlations} presents the rank correlations between the two composite metrics, grouped by combinations of training and inference strategies.

When analysing each strategy individually, the strongest correlation is observed for GRPO with Vanilla inference, with a Spearman's $\rho$ of 0.919 and Kendall's $\tau$ of 0.809. However, a notable trend appears when inference-time strategies are introduced, resulting in consistently lower correlation scores. This trend is also evident when grouping by the inference strategy, where vanilla inference shows the highest correlation regardless of the training strategy. When grouping by the training strategy, SFT has the highest correlation under Spearman's $\rho$ by a narrow margin, and GRPO demonstrates the highest correlation under Kendall's $\tau$.

These results highlight a notable practical phenomenon that while the generated code quality better aligns with the visual quality under SFT and GRPO, the enhancements made at the inference level prevent that correlation from holding. Instead, they produce code that differs from the expected but still results in videos that look similar to the anticipated output. This can also be seen as the models attempting to fix broken code with innovative but sub-optimal methods, which could lead to increasingly high-quality videos that closely resemble the expected video.

\begin{table}[t]
\centering
\caption{Rank correlations between Visual Similarity and CodeBERTBLEU, grouped by training and inference strategies. n denotes the number of model observations. n=17 for individual strategy combinations and n=51 (17*3) for grouped analyses}
\label{tbl:result-correlations}
\begin{tiny}
\begin{tabular}{@{}llccc@{}}
\toprule
\textbf{Grouping}                                                & \textbf{Category} & \textbf{Spearman's $\rho$} & \textbf{Kendall's $\tau$} & \textbf{n} \\ \midrule
\multirow{9}{*}{\textbf{Training $\times$ Inference Strategy}} & Base              & 0.689               & 0.559              & 17         \\
 & SFT         & 0.833          & 0.676          & 17 \\
 & GRPO        & \textbf{0.919} & \textbf{0.809} & 17 \\
 & Base+RITL     & 0.696          & 0.544          & 17 \\
 & SFT+RITL      & 0.799          & 0.603          & 17 \\
 & GRPO+RITL     & 0.713          & 0.559          & 17 \\
 & Base+RITL-DOC & 0.743          & 0.559          & 17 \\
 & SFT+RITL-DOC  & 0.694          & 0.544          & 17 \\
 & GRPO+RITL-DOC & 0.701          & 0.574          & 17 \\ \midrule
\multirow{3}{*}{\textbf{Training Strategy}}             & Base              & 0.703               & 0.527              & 51         \\
 & SFT         & \textbf{0.791} & 0.605          & 51 \\
 & GRPO    & 0.772          & \textbf{0.619} & 51 \\ \midrule
\multirow{3}{*}{\textbf{Inference Strategy}}            & Vanilla           & \textbf{0.817}      & \textbf{0.648}     & 51         \\
 & RITL          & 0.729          & 0.553          & 51 \\
 & RITL-DOC      & 0.706          & 0.534          & 51 \\ \bottomrule
\end{tabular}
\end{tiny}
\end{table}

\subsubsection{Qualitative Case Studies: Agreement of quantitative metrics and perceived quality}

\begin{figure}%
    \centering
    \subfloat[\centering GRPO outperforming the base version of the Qwen 3 Coder 30B model.]{{
    \includegraphics[width=0.4\linewidth]{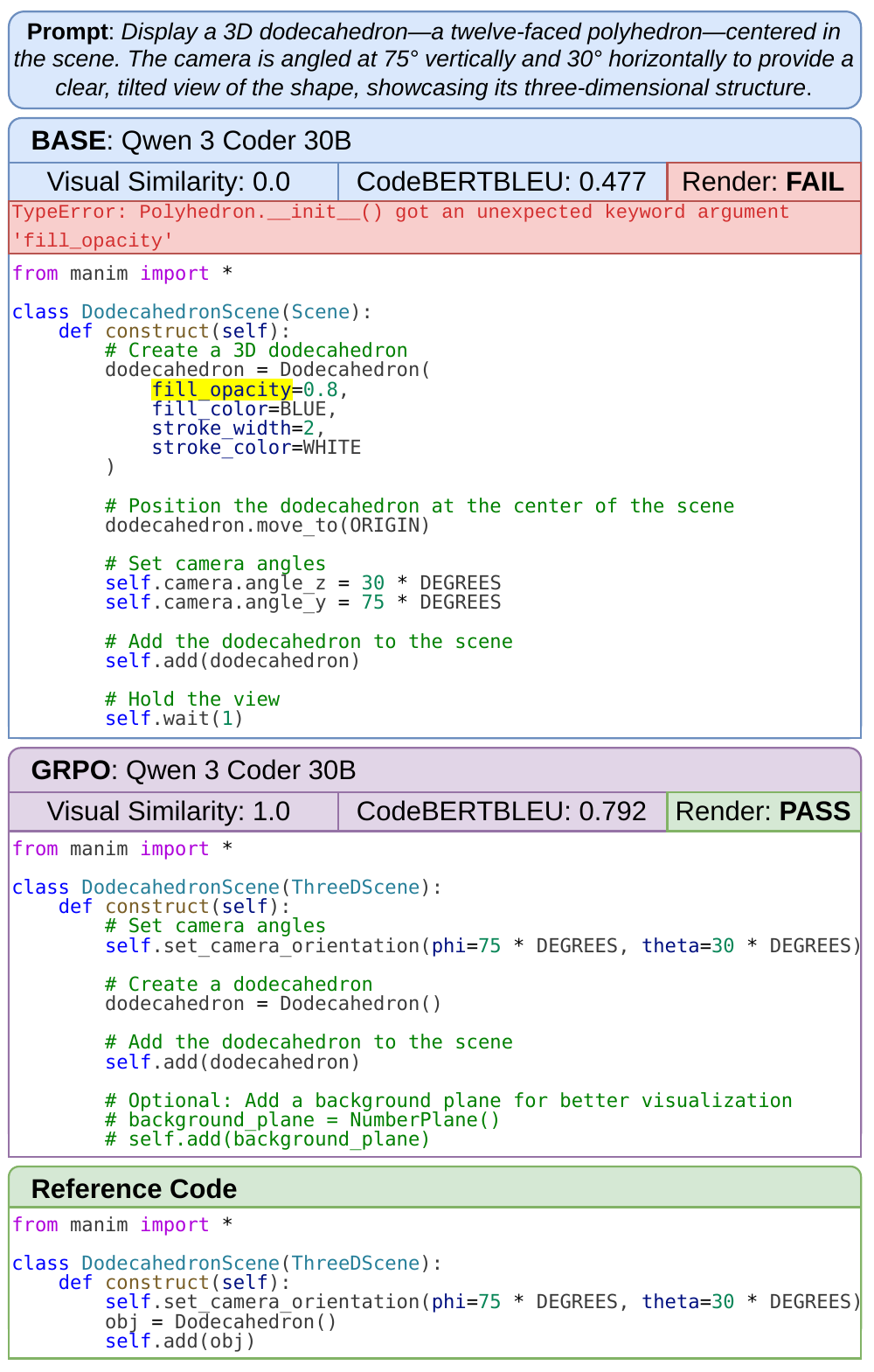}
    }}%
    \qquad
    \subfloat[\centering GRPO outperforming the base version of the Ministral 3 8B model.]{{
    \includegraphics[width=0.4\linewidth]{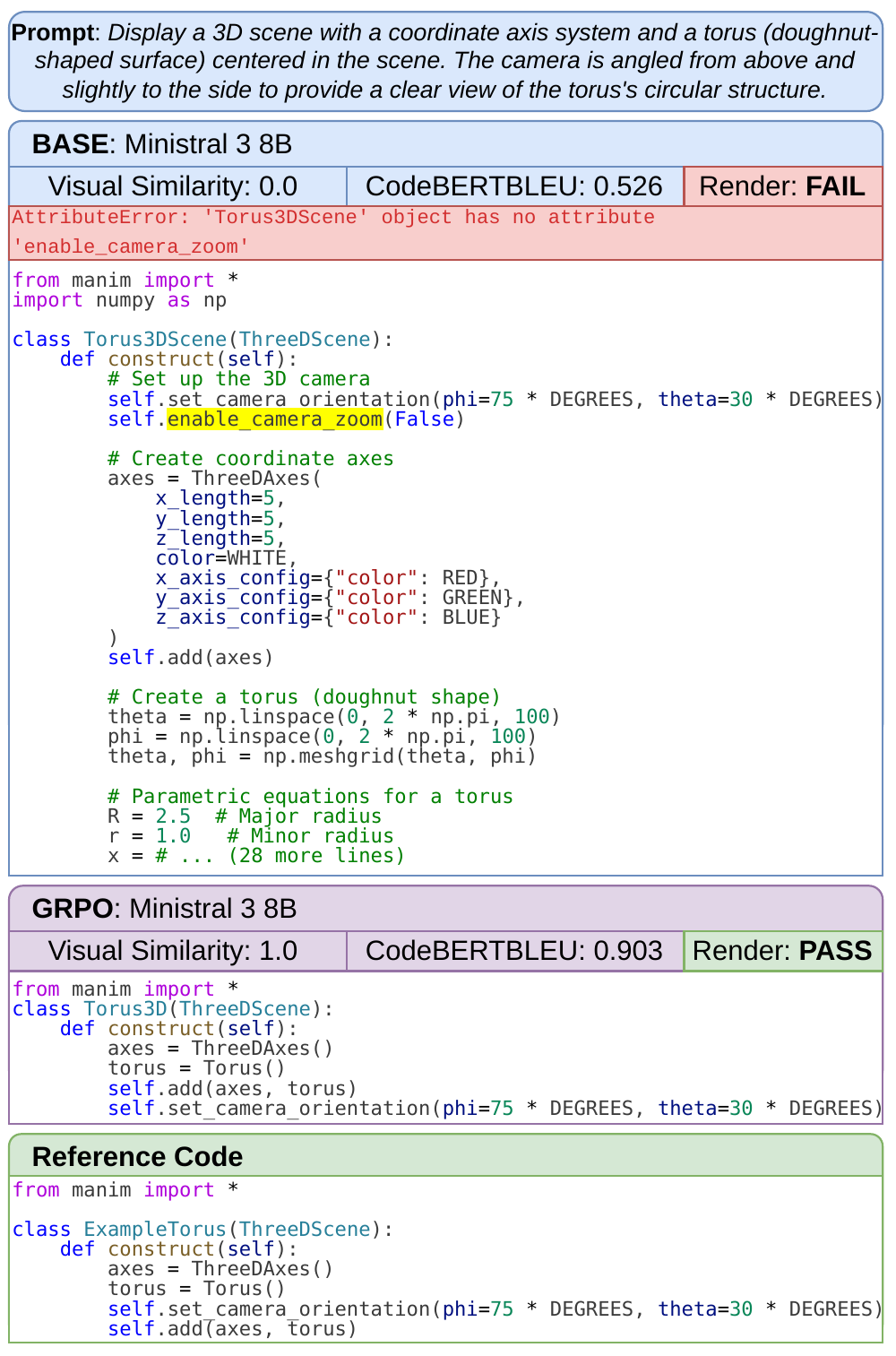}
    }}%
    \caption{Best Case: Fine-tuned model surpasses the base model.}%
    \label{fig:qualitative-bestcase}%
\end{figure}

\paragraph{Best Cases}
Figure~\ref{fig:qualitative-bestcase} shows two examples where GRPO fine-tuning improved the generated Manim code. The Qwen 3 Coder 30B example in Figure~\ref{fig:qualitative-bestcase} illustrates that the base model hallucinated API parameters, whereas GRPO guided the model to keep the default parameter value, as in the expected code. Similarly, in the Ministral 3 8B example, the base model hallucinates a built-in function when it is not even needed to create the scene described by the prompt. The GRPO model recognises this and produces code very similar to the reference code.

\begin{figure}%
    \centering
    \subfloat[\centering GRPO underperforming the base version of the LLaMA 3.2 3B model.]{{
    \includegraphics[width=0.4\linewidth]{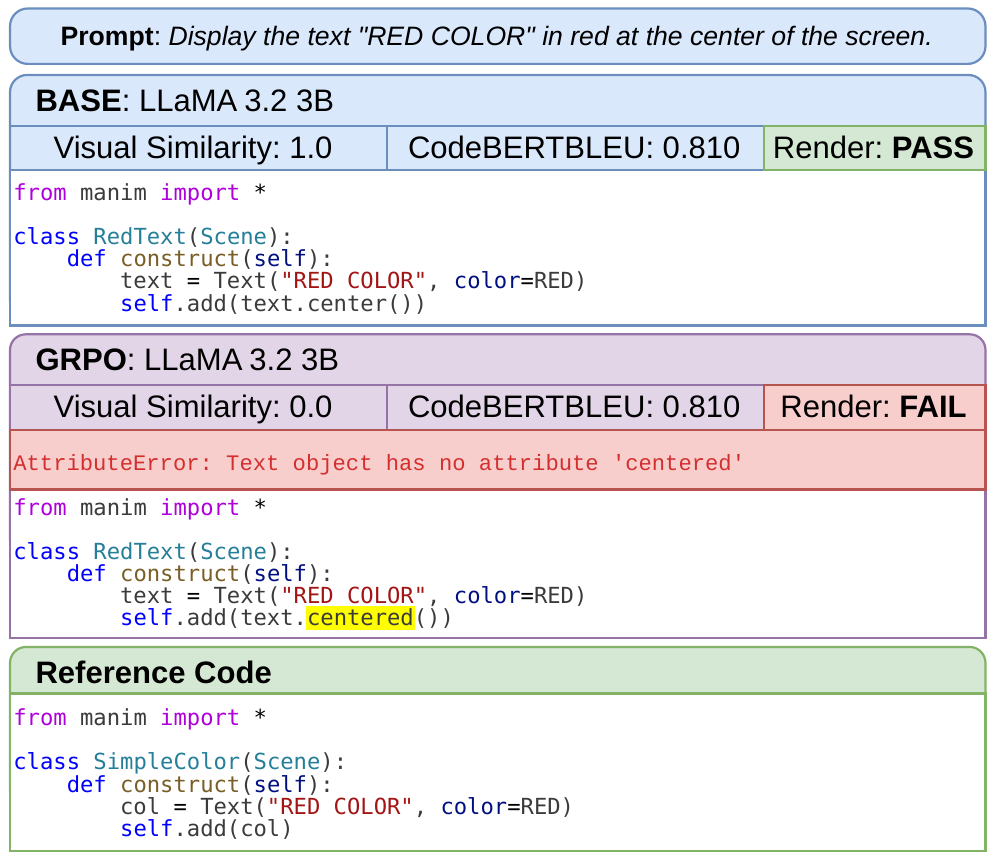}
    }}%
    \qquad
    \subfloat[\centering GRPO underperforming the base version of the Qwen 2.5 Coder 7B model.]{{
    \includegraphics[width=0.4\linewidth]{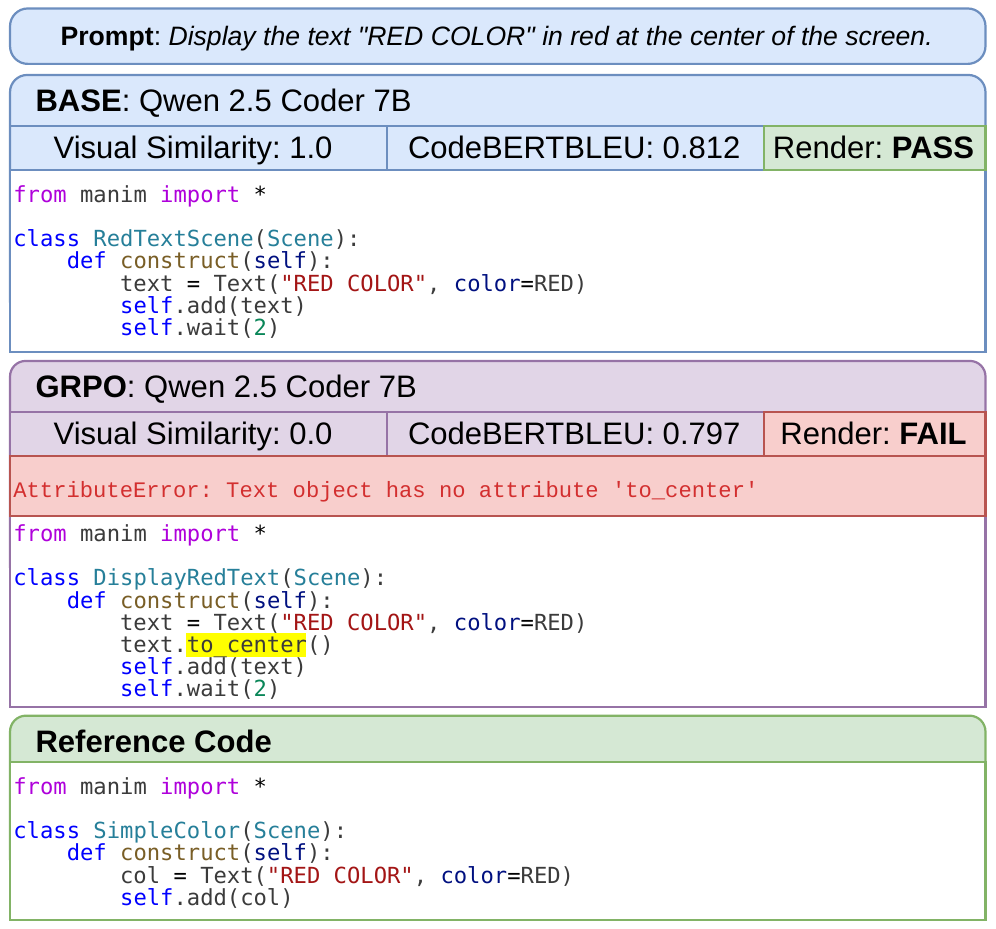}
    }}%
    \caption{Worst Case: Base model surpasses fine-tuned model}%
    \label{fig:qualitative-worstcase}%
\end{figure}

\paragraph{Worst Cases}
Figure~\ref{fig:qualitative-worstcase} illustrates failure cases where GRPO has caused the degradation of the performance of a base model. One similarity between these two examples is that both are from much smaller models with less than 8B parameters. The examples show that the increased spatial awareness of GRPO models has led them to hallucinate functions and API parameters that position visual elements in the required positions.

\begin{figure}%
    \centering
    \subfloat[\centering RITL inferencing outperforming Vanilla inferencing on SFT Qwen 3 Coder 30B model.]{{
    \includegraphics[width=0.4\linewidth]{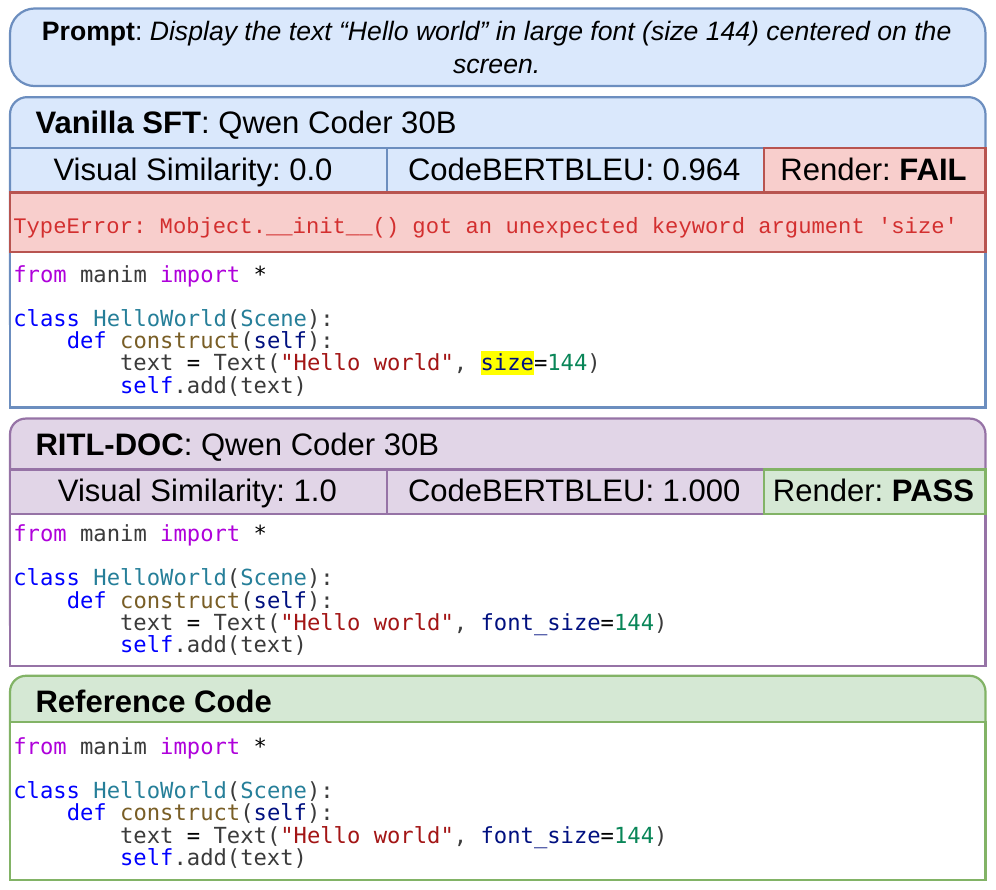}
    }}%
    \qquad
    \subfloat[\centering RITL inferencing outperforming Vanilla inferencing on SFT Qwen 3 Coder 30B model.]{{
    \includegraphics[width=0.4\linewidth]{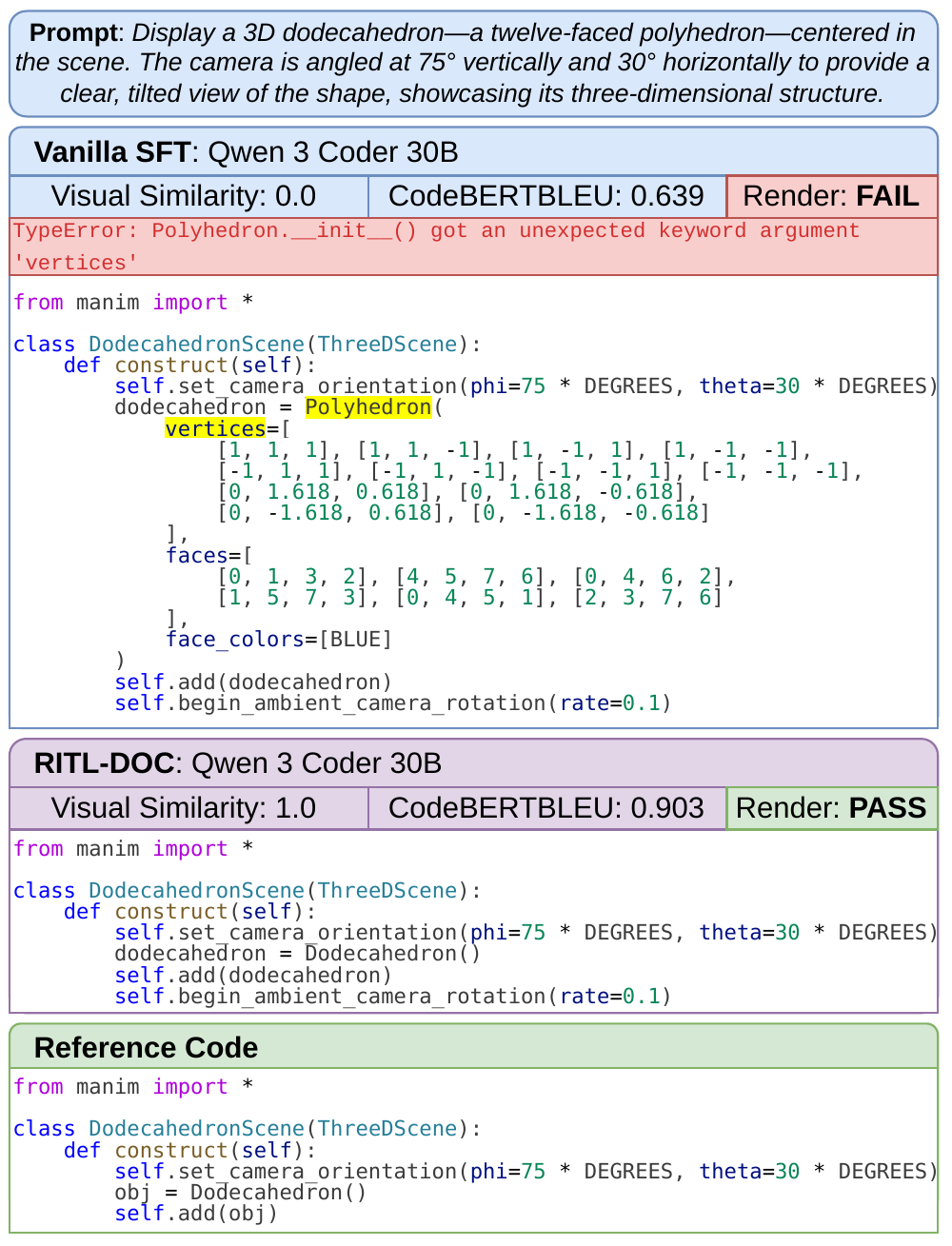}
    }}%
    \caption{Rescue Case: Inference strategy outperforming vanilla inference.}%
    \label{fig:qualitative-rescue}%
\end{figure}

\paragraph{Rescue Cases.}
Examples in Figure~\ref{fig:qualitative-rescue} demonstrate cases where inference-level strategies corrected erroneous code. The figure illustrates an instance of RITL-DOC fixing a hallucinated API parameter with the correct one using Qwen 3 Coder 30B. The right figure depicts the same model learning to correctly utilise the Manim API by employing the visual shapes available out of the box rather than attempting to recreate them.

\paragraph{Outlier Cases.}

\begin{figure}
    \centering
    \includegraphics[width=0.5\linewidth]{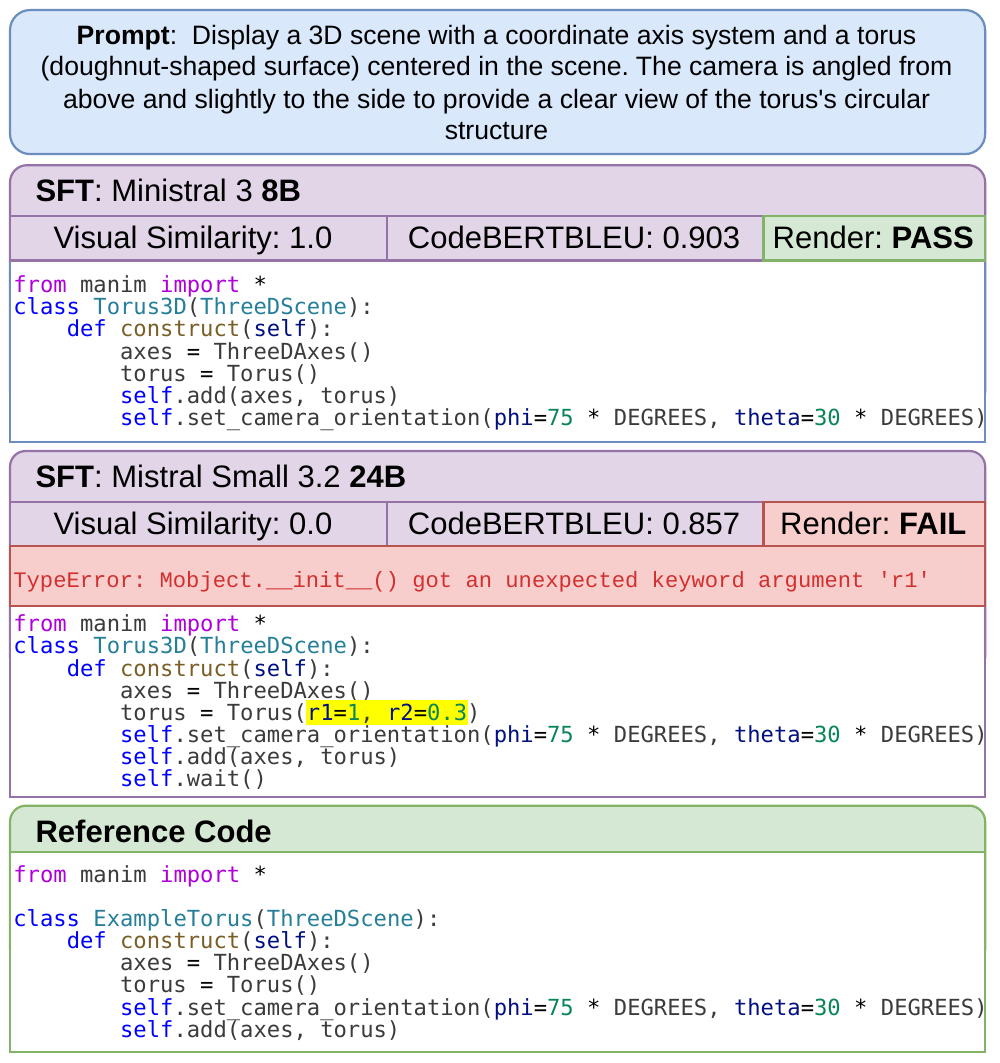}
    \caption{Outlier Case: An example of a smaller Ministral 8B SFT model outperforming Mistral Small 3.2 24B SFT model.}
    \label{fig:qualitative-outlier-1}
\end{figure}

Figure~\ref{fig:qualitative-outlier-1} shows an outlier where a smaller Ministral 3 8B SFT model successfully generated an error-free Manim code when the much larger Mistral Small 3.2 24B model failed. However, as seen in the figure, this is because the model tried to be precise about the visual object's parameters with hallucinated API parameters.

\paragraph{Visual Examples.}

\begin{figure}[t]
    \centering
    \includegraphics[width=1\linewidth]{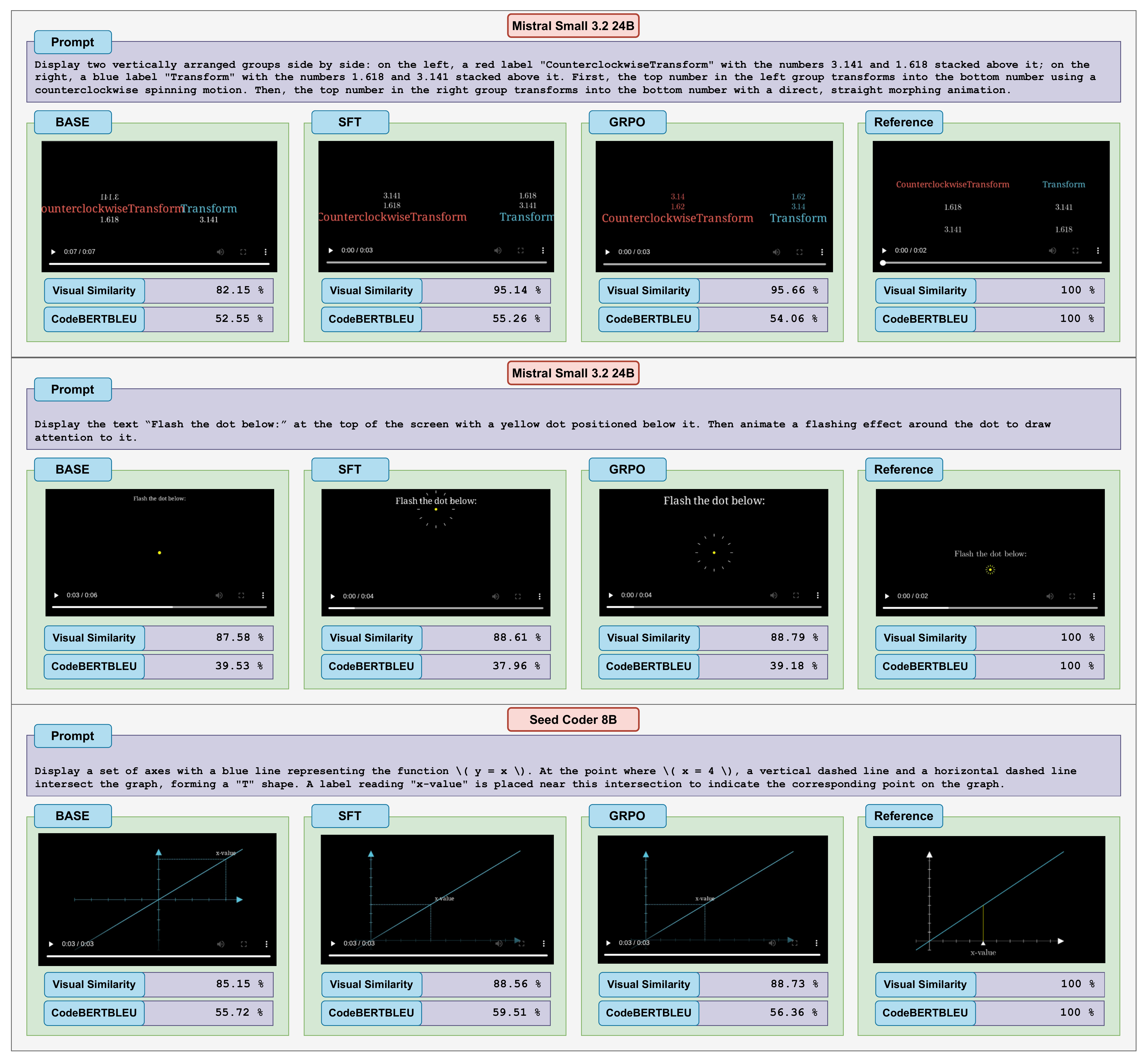}
    \caption{Examples from the top performing models. The top example shows that the GRPO has fixed the issue of text being out of the viewing area, while the middle example shows the GRPO fixing overlapping visual elements. The bottom example shows a case where the training cycles become increasingly similar to the reference video.}
    \label{fig:examples-stacked}
\end{figure}

Figure~\ref{fig:examples-stacked} shows a few selected visual examples where the training iterations have increased the visual quality and the accuracy of the generated videos.

\subsection{Limitations: Threats to Validity}

\paragraph{Internal Validity} 
Hyperparameter selection for both training strategies was conducted empirically and kept consistent across all models. Further hyperparameter optimisation tailored to individual models may produce different outcomes. Additionally, all models utilised LoRA adapters, which could introduce quantisation effects varying across models and model families.

\paragraph{External Validity} 
The experiments in this study mainly use ManimBench. While ManimBench covers the entire Manim API, it does not include samples that generate longer, script-based videos. Therefore, it is only suitable to draw conclusions about the fundamental capabilities of LLMs in Manim code generation, which could naturally extend to creating longer Manim videos. Additionally, the findings are specific to Manim Community Edition 0.19.0 and may not apply to other types of programmatic animation. Moreover, for all models, the study uses an Unsloth 4-bit quantised version due to resource limitations and because the focus was on evaluating compact models for Manim code generation. Full-precision models might theoretically achieve better performance but would require increased compute.

\paragraph{Construct Validity} 
The metrics used in the study aim to provide a holistic interpretation of a given LLM's capabilities for Manim code generation. However, the scores are based on the reference code, which is expert-written and from Manim's official documentation. This may penalise generated videos that are visually relevant to the provided textual description and high-quality, but differ from the reference code and video. Also, in error logs utilised in RITL, were truncated to the last 10 lines to reduce noise in the model context. While practically the last line of the error log contains the core reason for the error, it may occasionally truncate relevant diagnostic information.

\section{Conclusion and Future Work}
\label{sec:conclusion}
The paper presents ManimTrainer and ManimAgent, which together offer an end-to-end framework for training and deploying sub-30B LLMs to generate Manim programmatic animations. It provides the first comprehensive analysis of the interaction between training strategies and inference strategies for this task, using 17 open-source LLMs ranging from 0.5B to 30B parameters.

The results showed that both SFT and GRPO fine-tuning consistently improved the LLMs' Manim code generation ability, with SFT primarily enhancing code-level metrics and GRPO increasing the visual quality of the rendered videos, demonstrating their complementary roles. LLMs with around 8B parameters demonstrated well-balanced performance in terms of accuracy and efficiency, with the SeedCoder 8B model, trained with GRPO, outperforming the much larger Qwen 3 Coder Next 80B base model under vanilla inference. The model also performed comparably well to GPT-4.1.

Inference-time enhancement strategies achieved even greater performance gains than mere fine-tuning. Notably, GRPO models outperformed SFT and base models on RITL and RITL-DOC. The top overall results of 85.7\% VS and 94\% RSR were achieved by the Qwen 3 Coder 30B model with GRPO under RITL-DOC, utilising three RITL loops, and even surpassed the baseline GPT-4.1 model by +3.8 pp in VS and +2 pp in RSR. These findings confirm the hypothesis formulated at the beginning of this study that combining SFT with GRPO-based reinforcement learning and agentic inference strategies yields better performance in the LLM-based text-to-code-to-video process.

Analysis in this study indicates that code and visual metrics exhibit a relatively weak overall correlation, suggesting different coding approaches can produce similar visual outputs. Additionally, the evaluation results indicate that under vanilla inference, the code-visual metric correlation increased as the training pipeline progressed from the base to SFT, then from SFT to GRPO, while it consistently decreased when inference-level enhancements were introduced.

\paragraph{Future Work}
This work can be expanded in several directions. First, the frameworks presented can be used without modification to train much larger models in a multi-GPU setup, thereby validating these findings as parameter counts increase. Second, integrating a multimodal critic that offers natural-language feedback on the rendered videos could further address visual defects and inconsistencies. Third, the rule-based API retrieval in RITL-DOC can be improved by employing a learned retriever model to extract relevant API information, benefiting smaller models that currently find it difficult to handle longer, noisier contexts. Fourth, the efficiency of incorporating ManimAgent into a multimodal content generation system could be tested and assessed by human users. Finally, the ManimTrainer and ManimAgent frameworks could be adapted to other programmatic animation-generation domains, such as HTML/CSS, SVG-based animations, TikZ, and XML-based diagram generation, as they provide a highly efficient and precise alternative to image-generation models.

\newpage

\bibliographystyle{unsrt} 
\bibliography{bibliography}

\end{document}